\documentclass{article}
\usepackage[main,final]{neurips_2026}
\usepackage[utf8]{inputenc} 
\usepackage[T1]{fontenc}    
\usepackage{hyperref}       
\usepackage{url}            
\usepackage{booktabs}       
\usepackage{amsfonts}       
\usepackage{nicefrac}       
\usepackage{microtype}      
\usepackage{xcolor}         
\usepackage{amsmath,amssymb,amsthm,mathtools,bm,bbm}
\usepackage{mathrsfs}
\hypersetup{hidelinks}
\usepackage{enumitem}
\usepackage{geometry}
\usepackage{tikz}
\usepackage{algorithm}
\usepackage{algpseudocode}
\usepackage{graphicx}
\usetikzlibrary{positioning,arrows.meta,decorations.pathreplacing,calc}

\newcommand{\cS}{\mathcal{S}}
\newcommand{\cA}{\mathcal{A}}
\newcommand{\cY}{\mathcal{Y}}
\newcommand{\cZ}{\mathcal{Z}}
\newcommand{\cX}{\mathcal{X}}
\newcommand{\sH}{\mathsf{H}}
\newcommand{\E}{\mathbb{E}}
\newcommand{\Pp}{\mathbb{P}}
\newcommand{\ind}{\mathbbm{1}}

\newcommand{\logtwo}{\log_2}
\newcommand{\tv}[2]{\left\lVert #1-#2 \right\rVert_{\mathrm{TV}}}

\newcommand{\supp}{\mathrm{supp}}
\newtheorem{theorem}{Theorem}
\newtheorem{proposition}{Proposition}

\newtheorem{corollary}{Corollary}

\theoremstyle{remark}
\newtheorem{remark}{Remark}

\theoremstyle{definition}
\newtheorem{definition}{Definition}

\theoremstyle{remark}

\title{Finite-Memory Control of POMDPs: Fundamental Limits and Efficient Design}
\author{%
  Emrecan Kutay, Atilla Eryilmaz, Ness Shroff \\
  Department of Electrical and Computer Engineering\\
  The Ohio State University\\
  \texttt{kutay.5@osu.edu, eryilmaz.2@osu.edu, shroff.11@osu.edu}
}

\begin{document}

\maketitle

\begin{abstract}
Optimal control of partially observable Markov decision processes (POMDPs), in general, requires a complete history. The belief state, a distribution over hidden states, summarizes this history and provides the classical sufficient statistic to control POMDPs. However, the continuous nature of the belief space generally prevents a finite-memory controller from maintaining it exactly. This raises the question we address in this work: what must the available finite memory preserve to achieve near-optimal control performance? We study this question through an information-theoretic converse that gives a fundamental lower bound on the loss of any finite-memory controller. The analysis introduces the concept of a decision witness that characterizes belief states whose confusion would lead to unavoidable loss. Motivated by this bound, we propose \textsc{RECAP}, a constructive quantization method for designing representative memory states. We prove an upper bound on the optimality gap of any \textsc{RECAP} codebook and use the components of this bound together with the converse analysis to define the codebook objective. Experiments on 11 POMDP benchmarks demonstrate that \textsc{RECAP} improves over finite-memory representation baselines under the same memory budget.
\end{abstract}

\section{Introduction}
\label{sec:intro}
In a partially observable Markov decision process (POMDP), the agent receives only a partial signal about the true state. The full history becomes the sufficient statistic for future returns, because the Markov property no longer holds for observations. The belief state restores the Markov property by maintaining a distribution over states, but it lives on a continuous, high-dimensional simplex~\citep{SmallwoodSondik1973}. The full history may avoid this continuity, but it grows without bound over time, even in finite models~\citep{KaelblingLittmanCassandra1998}. Given these hard-to-maintain statistics, finite-state controllers (FSCs) seek the best way to operate under a hard memory budget, inherently discarding part of what the history encodes~\citep{PoupartBoutilier2003BFSC}.

In POMDPs, memory is a limited resource, and this paper studies its value: how much performance loss must a $B$-bit FSC incur solely because of its limited memory? We study this question from both directions: converse and construction. We first establish an information-theoretic converse that quantifies the unavoidable value loss of any $B$-bit FSC. We then introduce \textsc{RECAP}, a quantization design that constructs a finite memory guided by the belief distinctions that the converse analysis identifies as control-critical.

\paragraph{Related Work.}
The closest converse-adjacent work applies Fano's inequality to upper-bound the achievable reward from sensing information~\citep{MajumdarMeiPacelli2023SensorLimits}. It asks how informative the observations can be rather than how many bits of memory suffice. Recent work in~\citep{Nixon2026MyhillNerode} constructs a canonical lossless quotient of the history space by merging histories that no controller in a fixed test family can distinguish. Its target is the environment's structure, not the value cost of a lossy, budget-constrained memory. Learning-theoretic work on revealing POMDPs~\citep{ChenWangXiongMeiBai2023RevealingLB} establishes regret and sample-complexity bounds for online learning in an unknown environment, not the cost of a fixed finite-memory controller in a known model. None of these works establishes a converse for the memory bottleneck we study: the unavoidable value loss resulting from a hard $B$-bit constraint on any FSC.

On the constructive side, FSC synthesis methods search over controller structures, action mappings, and memory-update rules subject to value, budget, or specification constraints~\citep{PoupartBoutilier2003BFSC, WrayZilberstein2019GeneralizedControllers, AndriushchenkoEtAl2022InductiveFSC,SpelSteinKatoen2022OptimalBudget, WintererWimmerBeckerJansen2024StrongSimple, AndriushchenkoBorkCeskaJungesKatoenMacak2026Symbiotic}. They aim for constructive and provable policies, but their controller nodes encode what the search found useful, without any required relation to the belief, the history dynamics, or the future.

Approximate information-state and recurrent-summary methods study recursively updated summaries of the observation history~\citep{SubramanianSinhaSerajMahajan2022AIS, YangZhangAmiceLiTedrake2022DAIS,SinhaMahajan2024AgentState, SimaoSuilenJansen2023SPIPOMDP,LambrechtsBollandErnst2024InformedPOMDP}. The central condition across this literature is predictive sufficiency: the summary should preserve enough information about transition kernels, observation distributions, and reward that planning on it approximates planning on the full history.

Belief and value-directed compression methods focus on compressing the belief MDP in a value-aware manner~\citep{PoupartBoutilier2000VDBA,PoupartBoutilier2002VDC, RoyGordonThrun2005, LiHammarBertsekas2025FBBA,SaldiYukselLinder2020, KaraYuksel2022NearOptimality,CayciHeSrikant2024NACPOMDP}. Finite-model and finite-memory results characterize conditions under which discretized beliefs, aggregated features, or bounded histories preserve near-optimal value. The compressed object is assessed by the value loss it induces, not by any structural or predictive fidelity to the original representation. \textsc{RECAP} builds on all three lines, yet is not reducible to any of them. It neither searches over controller structure like FSC methods, nor certifies an existing summary as sufficient like information-state methods do, nor compresses the posterior along value-function directions as belief compression methods do. Instead, it proposes quantization design under a resource constraint, with updates following a predict-then-recompress recursion.

\paragraph{Contributions.} We study the fundamental limits of finite-memory control in discounted POMDPs. The constructive side of the problem has received substantial attention; what the converse side requires has not. Motivated by this, we investigate both directions and make the following contributions:

\begin{enumerate}[leftmargin=*]
    \item \textbf{Information-theoretic converse for POMDPs.} We prove in Theorem~\ref{thm:witness_lower_bound} an information-theoretic lower bound on the value loss of any $B$-bit FSC in a discounted POMDP. The argument introduces a ``decision witness'', an action-separating partition of belief space, which reduces finite-memory control to recovering a control-critical label from stored memory and the current observation. When $B$ bits are insufficient to resolve this label, Fano's inequality converts the resulting ambiguity into a concrete lower bound on value loss. We show that a witness-based bound is sharp in an example in Proposition~\ref{prop:noisy_tracking_lower_bound} and adapt the witness design to a different domain in Appendix~\ref{app:witness_example_ii}. Together, these examples demonstrate that decision witnesses provide useful bounds and are not tied to a specific model.

    \item \textbf{\textsc{RECAP}: a converse-motivated quantization design.} We introduce \textsc{RECAP}, a recursive action-aware predictive quantization method for budgeted FSC synthesis. Rather than compressing the posterior, \textsc{RECAP} maintains the predictive prior between observations. We prove in Theorem~\ref{thm:recap_regularized_bound} an upper bound on the optimality gap of the \textsc{RECAP} controller induced by any codebook and use the components of this bound together with the converse analysis to define the codebook objective that balances predictive coverage, recursive closure, and action regret.

    \item \textbf{Empirical validation.} We evaluate \textsc{RECAP} across 11 finite POMDP benchmarks. Simulations demonstrate that \textsc{RECAP} improves over baseline finite-memory representation methods under matched resource constraints (Table~\ref{tab:results}). This establishes the practical benefit of converse-driven recursive, action-aware predictive compression.
\end{enumerate}

\textit{Notation.} Unless stated otherwise, uppercase letters denote random variables and lowercase letters their values, except belief-valued random elements, which are written in lowercase. We use calligraphic letters for sets and alphabets; $\Pp,\E$ for probability and expectation; $H(\cdot)$, $I(\cdot;\cdot)$ for entropy and mutual information; $\Delta(\cX)$ for the simplex over $\cX$; and $\cdot\perp\cdot$ for independence. For $X^n=(X_1,\ldots,X_n)$, let $X_{1:t-1}$, $X_{t+1:n}$, and $X_{\setminus t}$ denote the past, future, and all but coordinate $t$, respectively. We write $[m:n]=\{m,\ldots,n\}$ and $[2^{nR}]=\{1,\ldots,2^{nR}\}$. $|v|_1$ is the $\ell_1$ norm, $\delta_a$ is a point mass, and $(\cdot)_+\triangleq \max \{0,\cdot\}$. Throughout, Bayes maps are completed by a fixed full-support fallback on zero-denominator observations.

\section{Control of POMDPs under Finite-Memory Constraints} 
We consider a discrete-time, time-homogeneous, finite-state POMDP with state space $\cS$, action space $\cA$, and observation space $\cY$, defined by
\begin{equation}
    e \triangleq (P_e,O_e,r_e,\xi_e,\gamma),
    \nonumber
\end{equation}
where $P_e(\cdot\mid s,a)\in\Delta(\cS)$ denotes the transition kernel, $O_e(\cdot\mid s)\in\Delta(\cY)$ the observation kernel, $r_e:\cS\times\cA\to[0,R_{\max}]$ the one-step reward function, $\xi_e\in\Delta(\cS)$ the initial law, and $\gamma \in (0,1)$ the discount. We write the history at time $t$ as $\sH_t \triangleq (Y_0,A_0,\ldots,Y_{t-1},A_{t-1},Y_t)$. For a full-history controller $\pi$, which selects the action $A_t$ based on $\sH_t$, we denote its discounted value by
\begin{equation}
    V_e^\pi \triangleq \E_e^\pi\!\left[\sum_{t=0}^\infty \gamma^t r_e(S_t,A_t)\right],
    \qquad 
    V_e^* \triangleq \sup_\pi V_e^\pi.
    \nonumber
\end{equation}
For a given history $\sH_t=h_t$, $b_t$ denotes the recursively updated belief,
\begin{equation}
    b_t(s) \triangleq \Pp_e(S_t=s\mid \sH_t=h_t)=\frac{\sum_{s'\in\cS} b_{t-1}(s')P_e(s\mid s',a_{t-1})O_e(y_{t}\mid s)}{\sum_{s',s''\in\cS} b_{t-1}(s')P_e(s''\mid s',a_{t-1})O_e(y_{t}\mid s'')},
    \qquad 
    s\in\cS.
    \nonumber
\end{equation}
The associated belief MDP has state space $\Delta(\cS)$, action space $\cA$, reward $r_e(b,a)=\sum_{s\in\cS} b(s)r_e(s,a)$, and transition kernel $\bar P_e(\cdot\mid b,a)= \sum_{y\in\cY} P_e(y\mid b,a)\delta_{\tau_e(b,a,y)}(\cdot)$, where $\tau_e(b,a,y)$ denotes the Bayes update. We write $V_e^*(b)$ and $Q_e^*(b,a)$ for the optimal value and action-value functions, respectively.

The belief ranges over the continuum $\Delta(\cS)$, which makes classical methods such as linear programming difficult to scale and their policies hard to realize with practical controllers. This motivates finite-memory summaries of history, whose design must preserve control-relevant information. One natural choice is a sliding window over the history, which is simple and model-free but allocates memory to the \emph{most recent} symbols rather than the \emph{most decision-relevant} ones. In contrast, a recursive statistic can preserve control-relevant cues across long periods of irrelevant observations. Appendix~\ref{app:motivating_counterexample} gives an example in which such a recursive statistic is optimal, whereas the sliding window under the same resource budget is not. We therefore work with the following general class of finite-state controllers, which contains both designs.

\begin{definition}[Finite-state controllers]
\label{def:fsc_main}
A $B$-bit finite-state controller is a policy with internal memory $Z_t\in\cZ$, $|\cZ|\le 2^B$. Let $\Pi_B$ denote the set of tuples
\begin{equation}
    \kappa=(\cZ,\nu_0,\{\pi_t\}_{t\ge0},\{\varphi_t\}_{t\ge0})
    \nonumber
\end{equation}
that induce
\begin{equation}
    Z_0\sim\nu_0(\cdot\mid Y_0),
    \qquad
    A_t\sim\pi_t(\cdot\mid Z_t,Y_t),
    \qquad
    Z_{t+1}\sim\varphi_t(\cdot\mid Z_t,Y_t,A_t).
    \nonumber
\end{equation}
For $\kappa\in\Pi_B$, define the budgeted value and memory-performance gap by
\begin{equation}
    V_e^\kappa \triangleq \E_e^\kappa\!\left[\sum_{t=0}^\infty \gamma^t r_e(S_t,A_t)\right],
    \qquad 
    V_e^*(B)\triangleq\sup_{\kappa\in\Pi_B}V_e^\kappa,
    \qquad 
    D_e(B)\triangleq V_e^*-V_e^*(B).
    \nonumber
\end{equation}
The quantity $D_e(B)$ is the smallest optimality gap under a $B$-bit memory constraint. Thus, $D_e(B)=0$ means that $2^B$ memory states suffice for optimal control, whereas $D_e(B)>0$ measures the unavoidable cost of finite memory. From an information-theoretic perspective, $Z_t$ is a lossy compression of the history $\sH_t$, so $D_e(B)$ describes a memory--performance tradeoff.
\end{definition}

In this work, we aim to develop controllers that employ efficient representations of belief states under their memory constraints. To this end, in the next two sections, we first lower bound $D_e(B)$ and then propose a nonuniform quantization scheme with an upper bound on its optimality gap.

\section{Memory-Performance Lower Bounds via Decision Witnesses}
\label{sec:converse}
In this section, we establish a lower bound on the memory-performance gap of a finite-memory controller for POMDPs. The idea behind our approach is the recognition that some regions of the belief space are more sensitive to finite-state representation than others, as measured by the suboptimality gap relative to the optimal actions associated with those beliefs. Accordingly, each partition of the belief space corresponds to a different lower bound that is based on: (i) how much uncertainty about the optimal action remains given the current observation and the memory, and (ii) how often the process enters these control-critical regions. Next, we formalize this idea by first introducing the concept of a \emph{decision witness}, and then establishing a fundamental lower bound on $D_e(B)$.

\begin{definition}[Decision witness at time $t$] 
\label{def:decision_witness}
Fix a time step $t\ge0$ and a number of labels $N\ge2$. Let $\psi:\Delta(\cS)\to\{0\}\cup[N]$ be a measurable labeling map, which partitions the belief space by assigning a label in $[N]$ or the null label $0$. For an observation subset $\cY_0\subseteq\cY$, an action selector $a^*:[N]\times\cY_0\to\cA$, and a margin $\eta>0$, define the witness label $\Theta_t\triangleq\psi(b_t)$ and the witness event $E_t\triangleq\{\Theta_t\in[N],\,Y_t\in\cY_0\}$. The tuple $(\psi,\cY_0,a^*,\eta)$ is a \emph{decision witness} at time $t$ if the following condition holds on $E_t$ at every value of $(b_t,Y_t)$ that occurs with positive probability under some controller:
\begin{equation}
    \eta\,\ind\{a\neq a^*(\Theta_t,Y_t)\} \le V_e^*(b_t)-Q_e^*(b_t,a), \qquad \forall a\in\cA,
    \nonumber
\end{equation}
and, for every $y\in\cY_0$, the map $i\mapsto a^*(i,y)$ is injective.
\end{definition}
The region indexed by the null label, $\Theta_t=0$, lies outside the ``witness event'' $E_t$. It contains beliefs for which either an alternative action may incur value loss below $\eta$, or the optimal action may be ambiguous. On the event $E_t$, however, the witness label is ``control-critical'', in that, for each observation, distinct labels in $[N]$ require distinct optimal actions, and choosing any other action loses at least $\eta$. Avoiding this loss therefore requires identifying $\Theta_t$ from the memory $Z_t$ and the current observation $Y_t$. This allows us to interpret the control problem as a classification problem over the $N$ distinct labels, where the information-theoretic approach uses Fano's inequality and yields the following lower bound on the optimality gap of every $B$-bit FSC.

\begin{theorem}[Decision witness lower bound]
\label{thm:witness_lower_bound}
Fix $t\ge0$, and let $(\psi,\cY_0,a^*,\eta)$ be a decision witness with $N$ labels at time $t$. Then every $B$-bit finite-state controller $\kappa\in\Pi_B$ satisfies
\begin{equation}
    \gamma^t \eta\,\Pp_e^\kappa(E_t)\,
    \inf_{\substack{P_{\Tilde Z_t\mid \sH_t}:\\ |\supp(\Tilde Z_t)|\le 2^B}}
    \left(\frac{H_e^\kappa(\Theta_t\mid Y_t,E_t)-I_e^\kappa(\Theta_t;\Tilde Z_t\mid Y_t,E_t)-1}{\log_2N}
    \right)_+
    \;\le\; V_e^*-V_e^\kappa \ .
\label{eq:witness_lower_bound}
\end{equation}
The left-hand side is zero when $\Pp_e^\kappa(E_t)=0$. The infimum ranges over all conditional laws of history summaries $\Tilde Z_t$ with alphabet size at most $2^B$. Consequently, $D_e(B)$ is lower-bounded by the infimum over $\kappa\in\Pi_B$ of the left-hand side of \eqref{eq:witness_lower_bound}.
\end{theorem}

\begin{proof}
See Appendix~\ref{app:witness_lower_bound}.
\end{proof}

\paragraph*{Interpretation of Theorem~\ref{thm:witness_lower_bound}.} Theorem~\ref{thm:witness_lower_bound} has three components.
\begin{itemize}
    \item \emph{Action gap.} The factor $\eta$ is the decision witness margin. On the witness event, every suboptimal action incurs a Bellman-value loss of at least $\eta$.

    \item \emph{Witness frequency.} The factor $\Pp_e^\kappa(E_t)$ is the probability, under controller $\kappa$, that the process enters a control-critical region at time $t$.

    \item \emph{Residual ambiguity under finite memory.} The inner term measures how much uncertainty about the active witness label remains after the current observation $Y_t$ and a $2^B$-state memory are available. The entropy term $H_e^\kappa(\Theta_t\mid Y_t,E_t)$ is the uncertainty about the active witness label after observing $Y_t$ on the witness event. The mutual information term $I_e^\kappa(\Theta_t;\Tilde Z_t\mid Y_t,E_t)$ is the amount of that uncertainty resolved by the memory $\Tilde Z_t$. The normalization by $\log_2 N$ expresses the remaining ambiguity as a fraction of the information needed to distinguish the $N$ witness labels.

    \item \emph{Role of the current observation.} The current observation is not part of the memory budget and is available at decision time. If $Y_t$ alone identifies the active witness label, then $H_e^\kappa(\Theta_t\mid Y_t,E_t)=0$, and no memory is needed. On the other extreme, if $Y_t$ gives no information about $\Theta_t$ on $E_t$, then the residual ambiguity depends entirely on the information provided by the memory $\Tilde Z_t$.
\end{itemize}

Consequently, finite memory incurs unavoidable value loss when the witness event has non-zero probability, the action gap is positive, and no $2^B$-state memory reliably identifies the active witness label after observing $Y_t$. Summing witnesses across time extends the one-step argument over a horizon. We state this extension as Corollary~\ref{cor:summed_lower_bound} in Appendix~\ref{app:witness_lower_bound}. The lower bound above quantifies the value loss for any fixed witness. Optimizing over decision witnesses yields the strongest bound within this framework. The example below instantiates the witness design and shows that the resulting lower bound can be tight with a matching FSC.

\paragraph{A tight decision witness lower bound.} We construct a decision witness for a finite tracking POMDP, derive its lower bound, and provide a matching finite-state controller, showing tightness. The hidden position $X$ starts uniformly on the cycle $\mathbb Z_K$ and evolves as
\begin{equation}
    X_0\sim \mathrm{Unif}(\mathbb Z_K),\qquad
    X_{t+1}=X_t+D_{t+1}\pmod K,\qquad
    D_{t+1}\sim \mathrm{Unif}\{-m,\ldots,m\},
    \nonumber
\end{equation}
where $m\ge1$ and $K\ge 2m+1$. At each time, the observation $Y_t$ is either a beacon ($\mathrm b$) or an odometry ($\mathrm o$) reading. A beacon reports the position $X_t$, and an odometry reading reports only the increment $D_t$, which locates $X_t$ relative to $X_{t-1}$. For $t\ge1$, the sensor returns a beacon with probability $q$ and an odometry reading with probability $1-q$. In either case, the reading is erased with probability $p$ and is reliable otherwise. Thus,
\begin{equation}
    Y_t\in \bigl(\{\mathrm b\}\times(\mathbb Z_K\cup\{\bot\})\bigr)
    \cup
    \bigl(\{\mathrm o\}\times(\{-m,\ldots,m\}\cup\{\bot\})\bigr),
    \nonumber
\end{equation}
where $Y_t=(\mathrm b,X_t)$ for a reliable beacon, $Y_t=(\mathrm o,D_t)$ for a reliable odometry reading, and $\bot$ denotes erasure. The first observation is a reliable beacon, $Y_0=(\mathrm b,X_0)$. The action is an estimate $a_t\in\mathbb Z_K$, and the reward is exact tracking $r(S_t,a_t)=\ind\{a_t=X_t\}$. Because the transition and observation kernels are action-independent, every controller induces the same law over the hidden trajectory and observations. Thus, memory affects performance only through the retained beacon and odometry information to estimate $X_t$. We provide a detailed description of the setting in Appendix~\ref{app:tracking_example}.

\begin{proposition}[Decision witness example lower bound]
\label{prop:noisy_tracking_lower_bound}
For the noisy beacon--odometry tracking model above, let $e=e(K,m,p,q)$ denote the corresponding environment. For each $t\ge1$, consider the decision witness with margin $\eta=1$ whose labels are the $K$ positions. Its witness event $E_t$ occurs when $Y_t$ and every reading since the last reliable beacon are reliable odometry readings, so that $b_t$ determines $X_t$ exactly. Summing these witnesses across time gives
\begin{equation}
    \sum_{t\ge 1}\gamma^t\,\alpha_t\left(1-\frac{2^B}{K}\right)_+\le D_e(B), \qquad
    \alpha_t \triangleq \rho^t+q(1-p)\sum_{t'=1}^{t-1}\rho^{t'}, \qquad
    \rho \triangleq (1-q)(1-p).
\label{eq:noisy_tracking_frontier}
\end{equation}
Moreover, if $p=0$ and $2^B\mid K$, this lower bound is tight: there exists a $B$-bit finite-state controller $\kappa_B^{\mathrm{qp}}\in\Pi_B$ such that
\begin{equation}
    V_e^*-V_e^{\kappa_B^{\mathrm{qp}}}
    =
    D_e(B)
    =
    \frac{\gamma(1-q)}{1-\gamma}\left(1-\frac{2^B}{K}\right).
    \label{eq:noisy_tracking_exact_frontier_p_zero}
\end{equation}
\end{proposition}

\begin{proof}
See Appendix~\ref{app:tracking_example}.
\end{proof}

Proposition~\ref{prop:noisy_tracking_lower_bound} shows how a decision witness yields a memory-performance lower bound, and the matching controller shows that this bound can be tight. The proof in Appendix~\ref{app:tracking_example} follows the steps of Theorem~\ref{thm:witness_lower_bound}, but bounds the classification error by a counting argument. This indicates that, while Theorem~\ref{thm:witness_lower_bound} uses Fano's inequality for its general applicability, the decision-witness framework admits sharper classification inequalities, which may depend on the setting, regime, or dynamics and yield tighter memory-performance lower bounds.

We provide another example in Appendix~\ref{app:witness_example_ii}, where a controller aims to identify the active hidden source from noisy observations; there, we design a decision witness for this setting and derive the corresponding lower bound. Together, these examples show that decision witnesses form a general framework that is not tied to a specific model. Instead, they provide an abstract way to expose belief distinctions that finite memory cannot safely erase. In the next section, we use this insight for a finite-memory design based on decision-aware predictive compression.

\section{\texorpdfstring
  {\textsc{RECAP}: A Design for \underline{\textbf{REC}}ursive \underline{\textbf{A}}ction-aware \underline{\textbf{P}}redictive Quantization}
  {\textsc{RECAP}: A Design for RECursive Action-aware Predictive Quantization}}

\begin{figure}[t]
\centering
\begin{tikzpicture}[
    scale=1,
    transform shape,
    >=Stealth,
    block/.style={
        draw,
        rounded corners,
        align=center,
        minimum height=0.68cm,
        minimum width=1.18cm,
        inner sep=2pt
    },
    wideblock/.style={
        draw,
        rounded corners,
        align=center,
        minimum height=0.68cm,
        minimum width=1.55cm,
        inner sep=2pt
    },
    arrow/.style={
        -{Stealth[length=2.1mm,width=1.45mm]},
        line width=0.5pt,
        shorten >=1.5pt,
        shorten <=1.5pt
    },
    faintarrow/.style={
        -{Stealth[length=1.6mm,width=1.1mm]},
        line width=0.4pt,
        gray!65,
        shorten >=1.5pt,
        shorten <=1.5pt
    },
    boundary/.style={line width=0.65pt},
    font=\scriptsize
]

\node[coordinate] (L) at (0,0) {};

\node[left=0.28cm of L] (zin) {$Z_t$};
\node[block, right=0.28cm of L] (rep) {$\bar c_{Z_t}$};

\node[wideblock, right=0.30cm of rep] (corr) {$\widehat b_t=C_{Y_t,\delta}(\bar c_{Z_t})$};
\node[block, right=0.30cm of corr] (act) {$A_t=\widehat a_H(\widehat b_t)$};
\node[block, right=0.30cm of act] (pred) {$\bar T_{A_t}(\widehat b_t)$};
\node[block, right=0.30cm of pred] (proj) {$\operatorname{enc}_{\mathcal C}$\\[-1pt]$(\cdot)$};

\node[coordinate, right=0.28cm of proj] (R) {};
\node[right=0.28cm of R] (zout) {$Z_{t+1}$};

\draw[boundary] ($(L)+(0,-0.76)$) -- ($(L)+(0,1.25)$);
\draw[boundary] ($(R)+(0,-0.76)$) -- ($(R)+(0,1.25)$);

\node[above=0.56cm of corr] (obs) {$Y_t\in\mathcal Y$};

\draw[arrow] (zin.east) -- (L);
\draw[arrow] (L) -- (rep.west);
\draw[arrow] (proj.east) -- (R);
\draw[arrow] (R) -- (zout.west);

\draw[arrow] (rep.east) -- (corr.west);
\draw[arrow] (corr.east) -- (act.west);
\draw[arrow] (act.east) -- (pred.west);
\draw[arrow] (pred.east) -- (proj.west);

\draw[arrow] (obs.south) -- (corr.north);

\coordinate (branch) at ($(corr.east)!0.5!(act.west)$);
\draw[arrow] (branch |- act.north) -- ++(0,0.38) -| (pred.north);

\node[below=0.08cm of rep]  {\tiny representative};
\node[below=0.08cm of corr] {\tiny correction};
\node[below=0.08cm of act]  {\tiny action};
\node[below=0.08cm of pred] {\tiny prediction};
\node[below=0.08cm of proj] {\tiny recompression};

\draw[faintarrow] ($(L)+(-0.94,0.62)$) -- ($(L)+(-0.48,0.62)$);
\node[align=center, above=1pt] at ($(L)+(-0.71,0.66)$) {\tiny from cycle\\[-1pt]\tiny $t-1$};

\draw[faintarrow] ($(R)+(0.48,0.62)$) -- ($(R)+(0.94,0.62)$);
\node[align=center, above=1pt] at ($(R)+(0.71,0.66)$) {\tiny to cycle\\[-1pt]\tiny $t+1$};

\node at ($(L)+(0,-1.12)$) {$t$};
\node at ($(R)+(0,-1.12)$) {$t+1$};

\draw[decorate,decoration={brace,mirror,amplitude=4pt}]
    ($(L)+(0,-0.86)$) -- ($(R)+(0,-0.86)$)
    node[midway,below=5pt] {\tiny time-$t$ control cycle};

\end{tikzpicture}
\caption{The control cycle at time $t$ of the \textsc{RECAP} controller in Definition~\ref{def:recap_grounded_structural}.}
\label{fig:recap_cycle}
\end{figure}
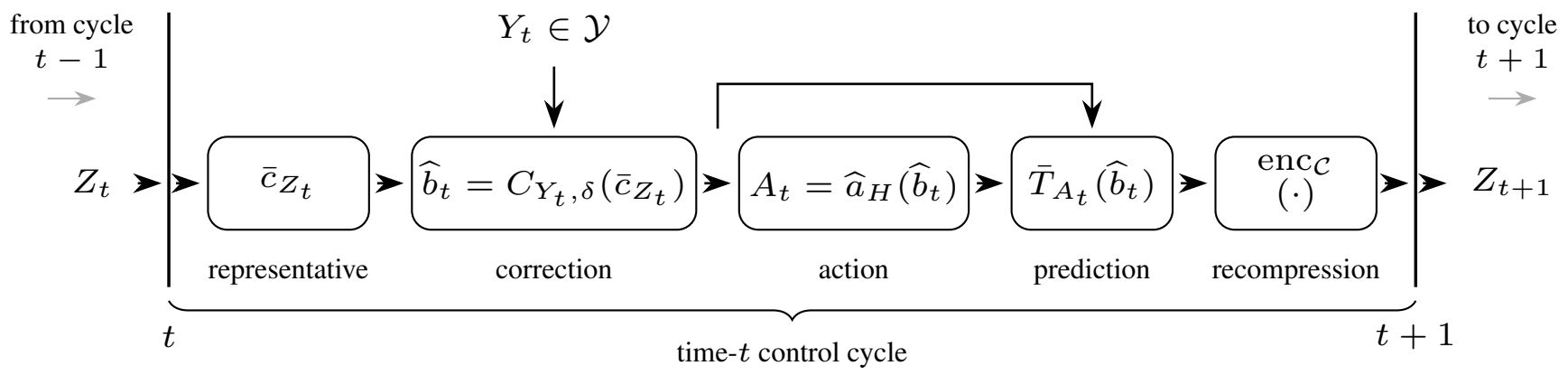

The converse results suggest that a good use of finite memory should preserve belief distinctions only when they matter for control. We follow this principle and introduce \textsc{RECAP}, a nonuniform quantizer for finite-state control. Given a memory budget $B$ and simulation rollouts, \textsc{RECAP} selects $K \le 2^B$ representative predictive priors, i.e., beliefs before the next observation arrives, from a pilot cloud of such priors generated by rollouts. These representatives form the codebook $\mathcal{C} = (\bar c_1,\ldots,\bar c_K)$ and serve as the internal states of a finite-state controller in Definition~\ref{def:fsc_main}. Figure~\ref{fig:recap_cycle} illustrates one \textsc{RECAP} control cycle, which builds on the prediction--correction decomposition of Bayesian filtering. Between control cycles, the controller stores one memory state $Z_t$. When a new observation $Y_t$ arrives, it corrects the stored representative $\bar c_{Z_t}$ using the observation, takes an action $A_t$ from the corrected belief $\widehat b_t$, propagates the result through the model dynamics $\bar T_{A_t}$, and recompresses to the next memory state $Z_{t+1}$. We can express these components mathematically by first defining the predictive prior as:
\begin{equation}
    \bar b_t(s)
    =
    \Pp_e(S_t=s\mid\sH_{t-1},A_{t-1})
    =
    \sum_{s'\in\cS}P_e(s\mid s',A_{t-1})b_{t-1}(s'),
    \qquad s\in\cS.
    \nonumber
\end{equation}
Then, we write the prediction step through the model dynamics as
\begin{equation}
    \nonumber
    \bar T_a(b)(s')
    \triangleq
    \sum_{s\in\cS}P_e(s'\mid s,a)b(s),
    \qquad
    \bar b_t
    =
    \bar T_{A_{t-1}}(b_{t-1}),
    \qquad
    t\ge1,
\end{equation}
with $\bar b_0\triangleq\xi_e$. After observing $Y_t=y_t$, the correction map is
\begin{equation}
    \nonumber
    C_{y_t}(\bar b_t)(s)
    \triangleq
    \frac{O_e(y_t\mid s)\bar b_t(s)}
    {\sum_{u\in\cS}O_e(y_t\mid u)\bar b_t(u)},
    \qquad s\in\cS.
\end{equation}
Finally, the exact filtering recursion decomposes as follows.
\begin{equation}
    b_t=C_{Y_t}(\bar b_t),
    \qquad
    \bar b_{t+1}=\bar T_{A_t}(b_t).
    \nonumber
\end{equation}

\subsection{\textsc{RECAP} Architecture}
\label{sec:recap_definition}
Following the architecture in Figure~\ref{fig:recap_cycle}, we define the \textsc{RECAP} controller induced by a predictive codebook $\mathcal{C}$ as follows.

\begin{definition}[\textsc{RECAP} architecture with parameter $\delta$]
\label{def:recap_grounded_structural}
Fix a finite POMDP model $e$, a memory budget $B$ as in Definition~\ref{def:fsc_main}, a parameter $\delta\in[0,1]$, and an $H$-step planner action rule $\widehat a_H:\Delta(\cS)\to\cA$. A \textsc{RECAP} controller with $K\le 2^B$ memory states is composed of the following components.

\begin{enumerate}[label=(\roman*),leftmargin=*]
\item \textbf{Predictive codebook.}
A codebook of representative predictive priors
\begin{equation}
    \nonumber
    \mathcal C=(\bar c_1,\ldots,\bar c_K),
    \qquad
    \bar c_j\in\Delta(\cS).
\end{equation}
The memory state $z\in\{1,\ldots,K\}$ means that the controller is carrying the predictive representative $\bar c_z$.

\item \textbf{Regularized correction.}
Fix a full-support anchor $\nu\in\Delta(\cS)$. For observation $y$, define
\begin{equation}
    \nonumber
    C_{y,\delta}(\bar b)(s)
    \triangleq
    \frac{
    O_e(y\mid s)\bigl((1-\delta)\bar b(s)+\delta\nu(s)\bigr)
    }{
    \sum_{u\in\cS}
    O_e(y\mid u)\bigl((1-\delta)\bar b(u)+\delta\nu(u)\bigr)
    },
    \qquad s\in\cS.
\end{equation}
\textsc{RECAP} corrects the stored predictive representative after first mixing it with a full-support anchor $\nu$. The parameter $\delta$ controls the amount of regularization and prevents degenerate Bayes denominators on feasible observations.

\item \textbf{Quantizer.}
The codebook induces a nearest-neighbor quantizer
\begin{equation}
    \operatorname{enc}_{\mathcal C}(\bar b)
    \in
    \operatorname*{argmin}_{j}
    \tv{\bar b}{\bar c_j},
    \label{eq:quantizer}
\end{equation}
with deterministic tie-breaking. Let $p_e(y\mid\bar b)\triangleq\sum_{s\in\cS}O_e(y\mid s)\bar b(s)$. The nearest-neighbor assignment is restricted to representatives $\bar c_j$ that preserve the feasible observations of $\bar b$, that is, $p_e(y\mid\bar c_j)>0$ whenever $p_e(y\mid\bar b)>0$, with a global nearest-neighbor fallback if no such representative exists.

\item \textbf{\textsc{RECAP} recursion.}
Given memory $Z_t$, \textsc{RECAP} applies the correction--action--prediction--compression cycle shown in Figure~\ref{fig:recap_cycle}. The initial memory is $Z_0= \operatorname{enc}_{\mathcal C}
(\xi_e)$, and the induced finite-state controller is
\begin{equation}
    A_t
    =
    \widehat a_H\!\left(C_{Y_t,\delta}(\bar c_{Z_t})\right),
    \qquad
    Z_{t+1}
    =
    \operatorname{enc}_{\mathcal C}\!\left(
    \bar T_{A_t}\!\left(C_{Y_t,\delta}(\bar c_{Z_t})\right)
    \right).
    \nonumber
\end{equation}
\end{enumerate}
\end{definition}

\begin{remark}[Predictive rather than posterior memory]
\textsc{RECAP} addresses an architectural question: under a hard FSC memory budget, what memory state should the controller carry between observations, refine when a new observation arrives, and update recursively after acting? \textsc{RECAP} answers by compressing predictive priors rather than posterior beliefs, departing from standard belief compression. This choice matches the timing in Definition~\ref{def:fsc_main}: for $t\ge1$, $Z_t$ is formed before $Y_t$ arrives, so it can carry information about $S_t$ only through the predictive prior $\bar b_t$, which $Y_t$ then corrects at decision time. This is the same split between memory and current observation that underlies Theorem~\ref{thm:witness_lower_bound}. We expect this predictive viewpoint to be natural when control and observation operate on different time scales, as in robotic or autonomous-driving systems, where a compact internal prediction must persist between slower sensor-fusion updates.
\end{remark}

We now bound the optimality gap of the \textsc{RECAP} controller $\kappa_{\mathcal C,\delta}$ induced by a codebook $\mathcal C$ and a regularization level $\delta\in(0,1]$. The performance bound accounts for three sources of error: finite-horizon planning, regularized correction, and quantization. The prediction step introduces no additional error because $\bar T_a$ is nonexpansive in TV distance under a common action.

The finite-horizon planner uses the optimal $H$-step truncated action-value function $\widehat Q_H$, which ignores rewards after the first $H$ steps. Under nonnegative rewards bounded by $R_{\max}$, the truncation error satisfies $0\le Q_e^*-\widehat Q_H\le\varepsilon_H\triangleq\gamma^HR_{\max}/(1-\gamma)$.

At the correction step, regularization introduces bias relative to exact Bayesian correction, while Bayesian correction can amplify predictive-prior errors when observation likelihoods are small. Specifically, $C_{y,\delta}$ deviates from exact Bayes correction $C_{y,0}$ by at most $\beta_\delta$ and, because mixing with the full-support anchor $\nu$ keeps its denominator at least $\delta\,p_e(y\mid\nu)$ for feasible observations, has TV sensitivity bounded by $\Lambda_\delta$, where
\begin{equation}
    \beta_\delta\triangleq\sup_{y,\,\bar b:\,p_e(y\mid\bar b)>0}\tv{C_{y,0}(\bar b)}{C_{y,\delta}(\bar b)},
    \qquad
    \Lambda_\delta\triangleq\max_{y:\,p_e(y\mid\nu)>0}\frac{2(1-\delta)}{\delta\,p_e(y\mid\nu)}.
    \label{eq:recap_correction_constants}
\end{equation}

At initialization and recompression, the quantizer in \eqref{eq:quantizer} incurs an error on the initial prior $\xi_e$ and on each predictive prior $\bar b^+_{z,y}\triangleq\bar T_{\widehat a_H(C_{y,\delta}(\bar c_z))}(C_{y,\delta}(\bar c_z))$ computed from $Z_t=z$ and $Y_t=y$:
\begin{equation}
    \bar e_{0,\mathcal C}\triangleq\tv{\xi_e}{\bar c_{\operatorname{enc}_{\mathcal C}(\xi_e)}},
    \qquad
    \bar\eta_{\mathcal C,\delta}\triangleq
    \sup_{z\in[K]}\sup_{y\in\cY}
    \tv{\bar b^+_{z,y}}{\bar c_{\operatorname{enc}_{\mathcal C}(\bar b^+_{z,y})}}.
    \label{eq:recap_codebook_errors}
\end{equation}
We combine these per-step errors to bound the performance gap of any \textsc{RECAP} codebook candidate as follows.

\begin{theorem}[\textsc{RECAP} performance bound]
\label{thm:recap_regularized_bound}
Fix a \textsc{RECAP} codebook $\mathcal C$ with $K\le 2^B$ and a regularization level $\delta\in(0,1]$ such that $\gamma\Lambda_\delta<1$, and let $\widehat a_H$ be greedy with respect to $\widehat Q_H$. Then the \textsc{RECAP} controller $\kappa_{\mathcal C,\delta}$ satisfies
\begin{equation}
    V_e^*-V_e^{\kappa_{\mathcal C,\delta}}
    \le
    \frac{2\varepsilon_H}{1-\gamma}
    +
    \frac{4R_{\max}}{1-\gamma}
    \left[
    \frac{\beta_\delta}{1-\gamma}
    +
    \Lambda_\delta
    \left(
    \frac{\bar e_{0,\mathcal C}}{1-\gamma\Lambda_\delta}
    +
    \frac{\gamma(\beta_\delta+\bar\eta_{\mathcal C,\delta})}
    {(1-\gamma)(1-\gamma\Lambda_\delta)}
    \right)
    \right].
    \label{eq:recap_regularized_bound}
\end{equation}
Since $\kappa_{\mathcal C,\delta}\in\Pi_B$, the right-hand side of \eqref{eq:recap_regularized_bound} also upper bounds $D_e(B)$. We derive an analogous bound for exact correction, $\delta=0$, under a local tube condition described in Appendix~\ref{app:recap_gap}.
\end{theorem}

\begin{proof}
See Appendix~\ref{app:recap_gap}.
\end{proof}

\paragraph*{Interpretation of Theorem~\ref{thm:recap_regularized_bound}.}
Theorem~\ref{thm:recap_regularized_bound} has five explicit ingredients:
\begin{itemize}
    \item \emph{Planner error.} The term $2\varepsilon_H/(1-\gamma)$ is the loss from using the finite-horizon planner instead of the optimal action-value function.

    \item \emph{Regularization bias.} The terms involving $\beta_\delta$ are the cost of using the regularized correction $C_{y,\delta}$ instead of exact Bayes correction $C_{y,0}$.

    \item \emph{Correction sensitivity.} The constant $\Lambda_\delta$ measures how much regularized correction can amplify predictive-prior error after an observation. The condition $\gamma\Lambda_\delta<1$ is a stability condition for the scalar error recursion.

    \item \emph{Initial predictive compression.} The term $\bar e_{0,\mathcal C}$ measures how well the initial predictive prior $\xi_e$ is represented by the codebook.

    \item \emph{Recursive reprojection.} The term $\bar\eta_{\mathcal C,\delta}$ measures how well the codebook represents the next predictive states produced by \textsc{RECAP}'s recursion. 
\end{itemize}

\subsection{\textsc{RECAP} Algorithm}
\label{sec:recap_algorithm}
The previous subsection defines the \textsc{RECAP} controller for a fixed predictive codebook $\mathcal C$. We now describe how \textsc{RECAP} constructs $\mathcal C$ from a given finite pilot cloud. Let
\begin{equation}
    \nonumber
    \mathcal{D}_{N_{\rm pilot}} \triangleq (\bar b_1, \ldots, \bar b_{N_{\rm pilot}}), \quad \bar b_i \in \Delta(\cS)
\end{equation}
be a given pilot predictive cloud obtained from simulation rollouts. For each $\bar b_i$, let $\cY_i$ be the set of feasible observations with observation probability $p_i(y)\triangleq p_e(y\mid\bar b_i)>0$. For $y\in\cY_i$, define 
\begin{equation}
    \nonumber
    b_{i,y} \triangleq C_{y,0}(\bar b_i).
\end{equation}

\paragraph{\textsc{RECAP} Objective.} 
Consider a candidate codebook $\mathcal C=(\bar c_1,\ldots,\bar c_K)$ with $K\le 2^B$, whose representatives are selected from the pilot cloud. Each pilot prior $\bar b_i$ is assigned to its representative,
\begin{equation} 
j_i\triangleq \operatorname{enc}_{\mathcal C}(\bar b_i), \quad \widehat b_{i,y}\triangleq C_{y,\delta}(\bar c_{j_i}), \quad \widehat a_{i,y}\triangleq \widehat a_H(\widehat b_{i,y}).
\nonumber
\end{equation}
The \textsc{RECAP} objective reflects both sides of the theory through three terms: predictive coverage and recursive closure, corresponding to the upper bound in Theorem~\ref{thm:recap_regularized_bound}, and action regret, which makes the codebook decision-aware as suggested by Theorem~\ref{thm:witness_lower_bound}. We begin with predictive coverage.
\begin{equation}
    J_{\rm pred}(\mathcal C)
    \triangleq
    \frac1{N_{\rm pilot}}\sum_{i=1}^{N_{\rm pilot}}
    \tv{\bar b_i}{\bar c_{j_i}}.
    \nonumber
\end{equation}
We next introduce recursive closure, which enforces stability under \textsc{RECAP}'s recursive updates.
\begin{equation}
    J_{\rm rec}(\mathcal C)
     \triangleq 
    \frac1{N_{\rm pilot}}
    \sum_{i=1}^{N_{\rm pilot}}
    \sum_{y\in\cY_i}
    p_i(y)
    \tv{
    \bar T_{\widehat a_{i,y}}(b_{i,y})
    }{
    \bar c_{
    \operatorname{enc}_{\mathcal C}\left(
    \bar T_{\widehat a_{i,y}}(\widehat b_{i,y})
    \right)}
    }.
    \nonumber
\end{equation}
The two TV-based terms above are natural for quantizing predictive distributions. Motivated by the converse results, we introduce the decision-aware action regret loss:
\begin{equation}
    J_{\rm act}(\mathcal C)
     \triangleq 
    \frac1{N_{\rm pilot}}
    \sum_{i=1}^{N_{\rm pilot}}
    \sum_{y\in\cY_i}
    p_i(y)
    \left[
    \max_{a\in\cA}\widehat Q_H(b_{i,y},a)
    -
    \widehat Q_H(b_{i,y},\widehat a_{i,y})
    \right].
    \nonumber
\end{equation}
This term incorporates the optimality gap incurred by using the representative-induced action instead of the original optimal action. If $b_{i,y}$ lies in a control-critical region, every suboptimal action loses at least the witness margin $\eta$ of Theorem~\ref{thm:witness_lower_bound}, and $J_{\rm act}$, a surrogate for this loss, penalizes codebooks that merge beliefs requiring different actions, reflecting the decision-witness condition in Definition~\ref{def:decision_witness}.

We add a penalty to prevent observation-incompatible assignments. Let $I_{\rm inc}(\mathcal C)$ be the number of pilot priors $\bar b_i$ whose assignment $j_i$ uses the fallback of the quantizer in~\eqref{eq:quantizer}. We set a penalty constant $\omega_{\rm inc} \gg 1$ and define
\begin{equation}
    P_{\rm inc}(\mathcal C)
     \triangleq 
    \omega_{\rm inc}\,\frac{I_{\rm inc}(\mathcal C)}{{N_{\rm pilot}}}.
    \nonumber
\end{equation}
Since $J_{\rm pred}$ and $J_{\rm rec}$ are total-variation losses while $J_{\rm act}$ has value units, we fix an action-loss scale $\zeta_{\rm act}>0$ and define the practical \textsc{RECAP} score with weights $\lambda_{\mathrm{pred}},\lambda_{\mathrm{rec}},\lambda_{\mathrm{act}}>0$ as
\begin{equation} 
    J_{\rm RECAP}(\mathcal C)
    = 
    \lambda_{\rm pred}J_{\rm pred}(\mathcal C) + 
    \lambda_{\rm rec}J_{\rm rec}(\mathcal C) + 
    \lambda_{\rm act}\frac{J_{\rm act}(\mathcal C)}{\zeta_{\rm act}} + 
    P_{\rm inc}(\mathcal C). 
    \label{eq:recap_objective}
\end{equation}

\textsc{RECAP} searches for a codebook with small $J_{\rm RECAP}$ by bounded medoid-style local search \citep{medoid}. It initializes $\mathcal C$ with farthest-first $k$-center in total variation, then performs coordinate-wise refinement. In each pass, the solver visits representatives sequentially. For representative $j$, it samples at most $m_{\rm cap}$ eligible pilot priors, evaluates replacements of $\bar c_j$, and immediately accepts the best strict improvement, so later representatives use the updated codebook. With $K$ representatives and pass limit $I_{\rm max}$, this gives $O(I_{\rm max}K m_{\rm cap})$ objective evaluations, unless refinement stops earlier after $p_{\rm stop}$ consecutive full passes with no accepted replacement. Algorithm~\ref{alg:RECAP} summarizes the procedure. For a fixed pilot cloud, planner, and candidate-sampling seed, each accepted replacement strictly decreases $J_{\rm RECAP}$, and the algorithm terminates after finitely many candidate evaluations.

\begin{algorithm}[t]
\footnotesize
\caption{\textsc{RECAP} codebook construction}
\label{alg:RECAP}
\begin{algorithmic}[1]
\State \textbf{Input:} pilot cloud $\mathcal D_{N_{\rm pilot}}$; budget $B$; parameter $\delta$; planner $\widehat Q_H$; objective $J_{\rm RECAP}$; pass limit $I_{\rm max}$; candidate cap $m_{\rm cap}$; patience $p_{\rm stop}$.
\State Set $K\leftarrow \min\{2^B,{N_{\rm pilot}}\}$, initialize $\mathcal C$ by farthest-first TV $k$-center, and set $n_{\rm stale}\leftarrow0$.
\For{$r=1,\ldots,I_{\rm max}$}
    \State $n_{\rm swap}\leftarrow0$.
    \For{$j=1,\ldots,K$}
        \State Sample at most $m_{\rm cap}$ eligible pilot priors $\bar b_i$ as replacements for $\bar c_j$.
        \State Let $\bar b_{i^\star}$ be the replacement with the smallest $J_{\rm RECAP}(\mathcal C^{j\leftarrow i})$.
        \If{$J_{\rm RECAP}(\mathcal C^{j\leftarrow i^\star})<J_{\rm RECAP}(\mathcal C)$}
            \State Replace $\bar c_j\leftarrow \bar b_{i^\star}$ and set $n_{\rm swap}\leftarrow n_{\rm swap}+1$.
        \EndIf
    \EndFor
    \State Set $n_{\rm stale}\leftarrow n_{\rm stale}+1$ if $n_{\rm swap}=0$, and $0$ otherwise.
    \If{$n_{\rm stale}\ge p_{\rm stop}$}
        \State \textbf{break}
    \EndIf
\EndFor
\State \Return codebook $\mathcal C$ and the induced \textsc{RECAP} controller from Definition~\ref{def:recap_grounded_structural}.
\end{algorithmic}
\end{algorithm}

\section{Experiments}
\label{sec:experiments}
We evaluate \textsc{RECAP} on 11 finite POMDP benchmarks from~\citep{pomdp_org}. All controllers operate under a hard $B$-bit memory budget. Due to space constraints, further experimental details appear in Appendix~\ref{app:app_experimental_details} whenever necessary. All experiments run on a single CPU machine (Intel Core i9-13950HX, 32 GB RAM) with no GPU acceleration. The complete simulation of 11 benchmarks across three design seeds requires approximately 25 CPU-hours in total, with per-benchmark runtimes ranging from 0.2 to 13 CPU-hours. The code is provided in the supplementary material and is also available at \url{https://github.com/RECAP-research/RECAP}.

\subsection{Experimental Setup}
\label{sec:experimental_setup}
We use the \textsc{RECAP} objective in \eqref{eq:recap_objective} with the same hyperparameter values across all benchmarks rather than instance-specific tuning: $\omega_{\rm inc}=10^6$, $\lambda_{\rm pred}=0.2$, $\lambda_{\rm rec}=0.3$, and $\lambda_{\rm act}=0.5$. We set the action-loss scale $\zeta_{\rm act}$ from the empirical $Q$-value spans over the pilot cloud. We initialize the codebook by farthest-first $k$-center in total variation distance, then run Algorithm~\ref{alg:RECAP} for up to $I_{\rm max}=16$ passes with at most $m_{\rm cap}=64$ swap candidates per representative, stopping early after $p_{\rm stop}=3$ consecutive passes without improvement. We fix the regularization level $\delta = 0.02$ with a uniform anchor $\nu$, and memory budgets $B$ are as in Table~\ref{tab:results}. We use $3$ design seeds per benchmark and evaluate each over $1000$ Monte Carlo episodes of length at most $128$. Benchmark descriptions, planner details, and pilot cloud sampling appear in Appendix~\ref{app:app_experimental_details}.

\subsection{Baselines}
\label{sec:baselines}
All baselines use the same memory budget and planner as \textsc{RECAP}, so performance differences solely reflect the memory representation. We compare against three classes: \textit{(i) Geometric predictive compression.} Predictive $k$-center places representatives by farthest-first coverage in total variation distance over the pilot cloud; predictive $k$-medoids minimizes empirical distortion over the same cloud. \textit{(ii) Feature-based compression.} Value-feature $k$-medoids clusters predictive priors by planner-induced action-value similarity; recursive-feature $k$-medoids clusters by one-step recursive branch features. \textit{(iii) Pilot-independent and history-based methods.} The Dirichlet simplex codebook draws representatives from the probability simplex; the finite observation-window baseline stores a bounded raw-observation suffix chosen to fit within $K$ states. We provide further details in Appendix~\ref{app:app_experimental_details}.

\renewcommand{\baselinestretch}{1.15}
\begin{table*}[t]
\centering
\small
\setlength{\tabcolsep}{5pt}
\caption{Mean $\pm$ std discounted return under a $B$-bit memory, pooled across episodes and design seeds. \textbf{\underline{Bold underline}} marks the best among budget-matched methods per row, with ties shared.}
\label{tab:results}
\resizebox{\textwidth}{!}{%
\begin{tabular}{@{}llcccccccc@{}}
\toprule
Benchmark & $B$
  & \textsc{RECAP}
  & Pred-$k$c
  & Pred-$k$m
  & Val-$k$m
  & Rec-$k$m
  & Dirichlet
  & Fin-win
  & Full bel. \\
\midrule
  paint
  & 2
  & \textbf{\underline{$\bm{3.30 \pm 1.81}$}} & $-1.04 \pm 2.15$ & \textbf{\underline{$\bm{3.30 \pm 1.81}$}} & $0.13 \pm 3.20$ & \textbf{\underline{$\bm{3.30 \pm 1.81}$}} & $-0.65 \pm 2.61$ & $0.41 \pm 1.69$ & $3.30 \pm 1.81$ \\
  marking
  & 2
  & \textbf{\underline{$\bm{2.73 \pm 0.55}$}} & $2.56 \pm 0.59$ & $2.56 \pm 0.60$ & $2.52 \pm 0.60$ & $2.33 \pm 0.60$ & $2.58 \pm 0.59$ & $2.56 \pm 0.59$ & $2.75 \pm 0.55$ \\
  saci\_s12
  & 2
  & \textbf{\underline{$\bm{14.65 \pm 7.81}$}} & \textbf{\underline{$\bm{14.65 \pm 7.81}$}} & \textbf{\underline{$\bm{14.65 \pm 7.81}$}} & \textbf{\underline{$\bm{14.65 \pm 7.81}$}} & \textbf{\underline{$\bm{14.65 \pm 7.81}$}} & \textbf{\underline{$\bm{14.65 \pm 7.81}$}} & $-4.22 \pm 9.67$ & $14.65 \pm 7.81$ \\
  line4\_2goals
  & 2
  & \textbf{\underline{$\bm{0.47 \pm 0.57}$}} & $0.45 \pm 0.56$ & \textbf{\underline{$\bm{0.47 \pm 0.58}$}} & $0.44 \pm 0.56$ & \textbf{\underline{$\bm{0.47 \pm 0.58}$}} & \textbf{\underline{$\bm{0.47 \pm 0.58}$}} & $0.46 \pm 0.57$ & $0.47 \pm 0.58$ \\
\midrule
  shuttle
  & 3
  & \textbf{\underline{$\bm{32.94 \pm 4.81}$}} & $24.16 \pm 6.48$ & $24.83 \pm 6.87$ & $28.23 \pm 6.42$ & $27.74 \pm 7.11$ & $26.57 \pm 3.11$ & $0.00 \pm 0.00$ & $32.94 \pm 4.81$ \\
  network
  & 3
  & $272.62 \pm 181.09$ & $265.77 \pm 178.11$ & \textbf{\underline{$\bm{275.60 \pm 180.91}$}} & $263.62 \pm 190.64$ & $255.02 \pm 182.79$ & $193.20 \pm 185.39$ & $-108.57 \pm 224.48$ & $290.10 \pm 178.11$ \\
  1d\_noisy
  & 3
  & $1.11 \pm 0.43$ & $1.10 \pm 0.44$ & $1.08 \pm 0.39$ & \textbf{\underline{$\bm{1.24 \pm 0.31}$}} & $1.02 \pm 0.48$ & $1.02 \pm 0.48$ & $1.02 \pm 0.48$ & $1.24 \pm 0.31$ \\
  query\_s2
  & 3
  & \textbf{\underline{$\bm{355.80 \pm 36.12}$}} & $346.60 \pm 35.56$ & $353.92 \pm 37.12$ & $349.66 \pm 36.51$ & $353.32 \pm 36.63$ & $355.43 \pm 35.98$ & $247.24 \pm 35.81$ & $356.43 \pm 36.60$ \\
\midrule
  mini\_hall2
  & 4
  & \textbf{\underline{$\bm{2.72 \pm 0.57}$}} & $0.43 \pm 0.68$ & $1.94 \pm 0.92$ & $0.40 \pm 0.67$ & \textbf{\underline{$\bm{2.72 \pm 0.57}$}} & $1.70 \pm 1.15$ & $0.19 \pm 0.48$ & $2.72 \pm 0.57$ \\
  4x4
  & 4
  & \textbf{\underline{$\bm{3.71 \pm 0.48}$}} & $0.67 \pm 1.05$ & $0.54 \pm 0.83$ & $1.37 \pm 1.69$ & $1.48 \pm 1.57$ & $0.22 \pm 0.48$ & $0.61 \pm 0.79$ & $3.73 \pm 0.49$ \\
  4x5x2
  & 4
  & \textbf{\underline{$\bm{0.60 \pm 0.79}$}} & $0.04 \pm 0.20$ & $0.09 \pm 0.31$ & $0.08 \pm 0.28$ & $0.07 \pm 0.27$ & $0.10 \pm 0.32$ & $0.04 \pm 0.20$ & $2.08 \pm 0.36$ \\
\bottomrule
\multicolumn{10}{@{}l}{\scriptsize
  Pred-$k$c: predictive $k$-center;\;
  Pred-$k$m: predictive $k$-medoids;\;
  Val-$k$m: value-feature $k$-medoids;\;
  Rec-$k$m: recursive-feature $k$-medoids;\;
  Fin-win: finite window;\;
  Full bel.: unconstrained full-belief planner reference (not budget-constrained).} \\
\end{tabular}
}
\end{table*}
\renewcommand{\baselinestretch}{1.0}

\subsection{Empirical Results}
\label{sec:empirical_results}
Table~\ref{tab:results} reports the mean $\pm$ pooled standard deviation of discounted returns across evaluation episodes and design seeds. Under the same memory budget, \textsc{RECAP} is the only method consistently competitive across all eleven benchmarks, leading or tying the best comparator on nine and ranking second on the remaining two. Its largest margins over the strongest comparator occur on 4x4, 4x5x2, and shuttle, whose observations carry little or aliased location information (Appendix Table~\ref{tab:benchmarks}), so, as in Theorem~\ref{thm:witness_lower_bound}, memory must resolve what the current observation cannot. There, predictive $k$-center, \textsc{RECAP}'s unrefined initialization, stays far below \textsc{RECAP}, and predictive $k$-medoids, which refines for predictive coverage alone, recovers little of the gap. The finite window, which keeps recent rather than decision-relevant observations, falls far below all other methods on saci\_s12, shuttle, network, and query\_s2, consistent with the counterexample in Appendix~\ref{app:motivating_counterexample}. Since the codebook baselines share \textsc{RECAP}'s predictive architecture and differ only in which belief distinctions their codebooks keep, the gap suggests the practical advantage of the converse-guided \textsc{RECAP} objective and codebook design.

We observe that the pooled deviations mainly reflect episode-level return variability rather than seed instability. For most method--benchmark pairs, seed-level variation in the mean return is below $20\%$ of the pooled deviation.  We also observe that refinement typically stops by the no-improvement criterion in Algorithm~\ref{alg:RECAP} before reaching $I_{\rm max}$, suggesting a modest search effort.

\section{Conclusion} 
In this paper, we have studied finite-memory control in POMDPs through an information-theoretic lens. We have introduced decision witnesses as a converse tool: any $B$-bit controller that cannot resolve an action-separating partition of belief space must incur value loss (Theorem~\ref{thm:witness_lower_bound}), and a witness-based lower bound on $D_e(B)$ can be tight (Proposition~\ref{prop:noisy_tracking_lower_bound}). We have proposed \textsc{RECAP} as the constructive counterpart, a quantization design whose memory states carry explicit predictive semantics aligned with the converse and whose optimality gap is bounded by the three error sources of its control cycle (Theorem~\ref{thm:recap_regularized_bound}). Our experiments support the main principle that control-relevant distinctions in belief space must survive any lossy compression of the history unless the current observation resolves them. A natural next step is tracing the full rate-distortion curve for a given POMDP: as $B$ grows, how does achievable value loss shrink, and how closely do practical constructions track the converse bound?

\begin{ack}
We thank the anonymous reviewers and the area chair for their thoughtful comments and helpful suggestions that improved the paper. This work has been supported in part by the Army Research Laboratory under Cooperative Agreement Number W911NF-23-2-0225, by the ARO Grant W911NF-24-1-0103, by the U.S. National Science Foundation under the grants: NSF AI Institute (AI-EDGE) 2112471, CNS-2312836, CNS-2225561, and CNS-2239677, and by the Office of Naval Research under grant N00014-24-1-2729. The views and conclusions contained in this document are those of the authors and should not be interpreted as representing the official policies, either expressed or implied, of the Army Research Laboratory or the U.S. Government. The U.S. Government is authorized to reproduce and distribute reprints for Government purposes notwithstanding any copyright notation herein.
\end{ack}

\bibliographystyle{unsrt}
\bibliography{references}


\appendix

\section{Motivating Counterexample}
\label{app:motivating_counterexample}
This section provides a motivating counterexample that shows memory design matters under a hard resource constraint. The example is a finite POMDP in which a recursively updated statistic is globally optimal, whereas two natural recent-window controllers under the same resource budget remain strictly suboptimal.

\paragraph*{[Model.]} Fix an integer $B\ge 3$ and parameters
\begin{equation}
    \varepsilon\in(0,\tfrac12),
    \qquad
    \lambda\in(0,1),
    \qquad
    q\in(\varepsilon,\tfrac12).
    \nonumber
\end{equation}
Set
\begin{equation}
    L\triangleq2^B-1.
    \nonumber
\end{equation}
Define alphabets
\begin{equation}
    \cS_{\mathrm{pkt}}\triangleq[L]\times\{0,1\},
    \qquad
    \cY_{\mathrm{pkt}}\triangleq[L]\times\{0,1\},
    \qquad
    \cA_{\mathrm{pkt}}\triangleq\{0,1\}.
    \nonumber
\end{equation}
The hidden state is
\begin{equation}
    \nonumber
    S_t=(G_t,\Theta_t)\in\cS_{\mathrm{pkt}},
\end{equation}
where $G_t\in[L]$ is the current packet age and $\Theta_t\in\{0,1\}$ is the current packet label. Initialize at time $t=0$ with
\begin{equation}
\nonumber
    G_0=1,
    \qquad
    \Theta_0\sim \mathrm{Bernoulli}(q).
\end{equation}
For every $t\ge 0$, the next state is drawn as follows.
\begin{equation}
    \nonumber
    (G_{t+1},\Theta_{t+1})=
    \begin{cases}
        (1,\Theta'_{t+1}), \quad \Theta'_{t+1}\sim \mathrm{Bernoulli}(q),
        & G_t<L \text{ and refresh with prob. } \lambda,\\
        (G_t+1,\Theta_t),
        & G_t<L \text{ and no refresh with prob. } 1-\lambda,\\
        (1,\Theta'_{t+1}), \quad \Theta'_{t+1}\sim \mathrm{Bernoulli}(q),
        & G_t=L.
    \end{cases}
\end{equation}
Thus, if $G_t<L$, the packet label is redrawn and the age resets with probability $\lambda$. If $G_t=L$, a forced reset occurs. The state transition kernel is action-independent. At time $t$, the controller observes
\begin{equation}
    \nonumber
    Y_t=(G_t,R_t)\in\cY_{\mathrm{pkt}},
\end{equation}
where $R_t\in\{0,1\}$ is a noisy status bit satisfying
\begin{equation}
    \nonumber
    \Pp(R_t=1\mid \Theta_t=1)=1-\varepsilon,
    \qquad
    \Pp(R_t=1\mid \Theta_t=0)=\varepsilon.
\end{equation}
The controller chooses an action
\begin{equation}
    \nonumber
    A_t\in\cA_{\mathrm{pkt}}
\end{equation}
and receives the reward
\begin{equation}
    \nonumber
    r(S_t,A_t)=\ind\{A_t=\Theta_t\}.
\end{equation}
Because both the state transition and the observation kernels are action-independent, every policy induces the same marginal law of $(S_{0:t},Y_{0:t})$. For a policy $\pi$, define
\begin{equation}
    V_e^\pi=
    \E^\pi\!\left[\sum_{t=0}^\infty \gamma^{t}r(S_t,A_t)\right],
    \qquad \gamma\in(0,1).
    \nonumber
\end{equation}
Let
\begin{equation}
    \nonumber
    C_t=\sum_{j=t-G_t+1}^t R_j
\end{equation}
be the number of observed $1$'s accumulated since the last redraw. Then $0\le C_t\le G_t\le L$. The memory $Z_t$ of the recursive-count controller stores the previous count of observed $1$'s, with $Z_0=0$. At decision time, the controller recovers the current count from $(Z_t,Y_t)$ by
\begin{equation}
    \nonumber
    C_t=
    \begin{cases}
        \ind\{R_t=1\}, & G_t=1,\\
        Z_t+\ind\{R_t=1\}, & G_t>1.
    \end{cases}
\end{equation}
After acting, the controller updates its memory by
\begin{equation}
    \nonumber
    Z_{t+1}=C_t.
\end{equation}
Hence, the recursive-count controller uses $|\cZ|=L+1=2^B$ memory states. Let $\pi^{\mathrm{cnt}}\in\Pi_B$ denote the controller that, at time $t$, chooses the maximum a posteriori action for $\Theta_t$ from $(G_t,C_t)$:
\begin{equation}
    \nonumber
    A_t^{\mathrm{cnt}}
    =
    \arg\max_{u\in\{0,1\}} \Pp(\Theta_t=u\mid G_t,C_t).
\end{equation}

\emph{Recent status-bit window.}
Let $\mathcal W_B^{\mathrm{obs}}$ be the class of possibly randomized policies whose time-$t$ action kernel is measurable with respect to
\begin{equation}
    \nonumber
    (G_t,R_t,R_{t-B:t-1})
\end{equation}
together with exogenous controller randomization independent of the environment.

\emph{Recent observation--action window baseline (uses $2\lfloor B/2\rfloor$ retained bits).}
Let
\begin{equation}
    \nonumber
    n_B \triangleq \left\lfloor \frac{B}{2}\right\rfloor,
\end{equation}
and let $\mathcal W_B^{\mathrm{oa}}$ be the class of possibly randomized policies whose time-$t$ action kernel is measurable with respect to
\begin{equation}
\nonumber
(G_t,R_t,R_{t-n_B:t-1},A_{t-n_B:t-1})
\end{equation}
together with exogenous controller randomization independent of the environment. Because the state transition and observation kernels are action-independent, past actions reveal no hidden-state information beyond the observation history. Thus, although the observation--action class $\mathcal W_B^{\mathrm{oa}}$ is the convenient recent-window baseline in the literature, we also include the status-bit window class $\mathcal W_B^{\mathrm{obs}}$, which allocates the entire retained-memory budget to past observations. For this counterexample, $\mathcal W_B^{\mathrm{obs}}$ serves as a complementary observation-only recent-window baseline.

\begin{proposition}[Strict separation between memory designs]
\label{prop:packet_same_budget_gap}
Consider the packet-age model above with $B\ge 3$, $L=2^B-1$, $\varepsilon\in(0,\tfrac12)$, $\lambda\in(0,1)$, and $q\in(\varepsilon,\tfrac12)$. The recursive-count controller $\pi^{\mathrm{cnt}}\in\Pi_B$ uses exactly the $B$-bit memory budget and is globally optimal. It strictly outperforms both $B$-bit recent-window baselines:
\begin{equation}
    V_e^{\pi^{\mathrm{cnt}}}
    >
    \sup_{\pi\in\mathcal W_B^{\mathrm{obs}}} V_e^\pi,
    \qquad
    V_e^{\pi^{\mathrm{cnt}}}
    >
    \sup_{\pi\in\mathcal W_B^{\mathrm{oa}}} V_e^\pi,
    \qquad \forall\,\gamma\in(0,1).
\end{equation}
\end{proposition}

\begin{proof}
We first identify the maximum a posteriori rule induced by $(G_t,C_t)$ and use it to prove global optimality of $\pi^{\mathrm{cnt}}$. We then construct, for each recent-window class, two feasible histories with different optimal actions that no globally optimal policy in the class can distinguish, and use an attainment argument to conclude strict suboptimality.

Because the packet process is exogenous, actions do not affect either the hidden evolution or the observation law. Let
\begin{equation}
    \mathscr Y_t \triangleq \sigma(Y_{0:t})
    \nonumber
\end{equation}
denote the observation history up to time $t$. Because the packet dynamics and observations are action-independent, past actions reveal no information about $\Theta_t$ beyond the observation history that generates them. Thus, for any possibly randomized policy $\pi$,
\begin{equation}
    \nonumber
    \Pp^\pi(A_t=\Theta_t\mid \sH_t)
    =
    \sum_{u\in\{0,1\}} \pi_t(u\mid \sH_t)\,\Pp(\Theta_t=u\mid Y_{0:t})
    \le
    \max_{u\in\{0,1\}}\Pp(\Theta_t=u\mid Y_{0:t}).
\end{equation}
Taking expectations and summing over $t$ yields
\begin{equation}
    V_e^\pi
    \le
    \sum_{t\ge 0}\gamma^{t}
    \E\!\left[\max_{u\in\{0,1\}}\Pp(\Theta_t=u\mid Y_{0:t})\right].
    \label{eq:packet_universal_upper_bound}
\end{equation}
Conditioned on the event $G_t=a$ there are $a$ status bits since redraw
\begin{equation}
    \nonumber
    R_{t-a+1},\ldots,R_t.
\end{equation}
Under $\Theta_t=0$, these bits are i.i.d.\ Bernoulli$(\varepsilon)$. Under $\Theta_t=1$, they are i.i.d.\ Bernoulli$(1-\varepsilon)$. Hence, the posterior depends on the current packet's observation record only through
\begin{equation}
    C_t=\sum_{j=t-a+1}^t R_j.
    \nonumber
\end{equation}
For age $a\in\{1,\ldots,L\}$ and count $c\in\{0,\ldots,a\}$, Bayes' rule gives
\begin{equation}
    \nonumber
    \frac{\Pp(\Theta_t=1\mid G_t=a,C_t=c)}
         {\Pp(\Theta_t=0\mid G_t=a,C_t=c)}
    =
    \frac{q}{1-q}\left(\frac{1-\varepsilon}{\varepsilon}\right)^{2c-a}.
\end{equation}
Define
\begin{equation}
    \nonumber
    \Lambda \triangleq \log\frac{1-\varepsilon}{\varepsilon}>0,
    \qquad
    \alpha \triangleq \frac{1}{2\Lambda}\log\frac{1-q}{q}.
\end{equation}
Since $q\in(\varepsilon,\tfrac12)$, we have $0<\alpha<0.5$. Hence, the MAP action is
\begin{equation}
    \nonumber
    A_t^{\mathrm{cnt}}=
    \begin{cases}
        1, & C_t>\dfrac{G_t}{2}+\alpha,\\[0.6em]
        0, & C_t<\dfrac{G_t}{2}+\alpha.
    \end{cases}
\end{equation}
Since $C_t$ is an integer and $G_t/2+\alpha$ is not, ties never occur. Thus $\pi^{\mathrm{cnt}}$ chooses the unique maximum a posteriori action at every feasible history. Hence, it attains \eqref{eq:packet_universal_upper_bound} and is globally optimal. Since ties never occur in this model, every globally optimal policy must choose the unique maximum a posteriori action almost surely at every feasible history. We next construct two feasible age-$L$ histories that agree on the recent window information but have different current unique MAP actions. Fix
\begin{equation}
    \nonumber
    n\in\mathbb N_0
    \qquad\text{such that}\qquad
    n+1<\frac{L}{2}+\alpha.
\end{equation}
Define two age-$L$ paths
\begin{equation}
    \nonumber
    r_n^- \triangleq (0^{\,L-n-1},1^{\,n+1}),
    \qquad
    r_n^+ \triangleq (0^{\,(L-1)/2},1^{\,(L+1)/2}).
\end{equation}
They share the same last $n+1$ status bits, namely $1^{\,n+1}$, but have different total counts:
\begin{equation}
    \nonumber
    C(r_n^-)=n+1,
    \qquad
    C(r_n^+)=\frac{L+1}{2}.
\end{equation}
Since $0 < \alpha < 0.5$,
\begin{equation}
    \nonumber
    n+1<\frac{L}{2}+\alpha<\frac{L+1}{2}.
\end{equation}
Thus, the current unique MAP action is $0$ on $r_n^-$ and $1$ on $r_n^+$. Now examine the previous $n$ times, whose ages are $L-n$, $L-n+1$, $\ldots$, $L-1$. For $r_n^-$ the running counts are
\begin{equation}
    \nonumber
    1,2,\ldots,n,
\end{equation}
all strictly below the corresponding thresholds $a/2+\alpha$. For $r_n^+$ the running counts are
\begin{equation}
    \nonumber
    \frac{L+1}{2}-n,\;\frac{L+1}{2}-n+1,\;\ldots,\;\frac{L-1}{2},
\end{equation}
and at age $L-n+j$ the gap to the threshold is
\begin{equation}
    \nonumber
    \left(\frac{L-n+j}{2}+\alpha\right)
    -
    \left(\frac{L+1}{2}-n+j\right)
    =
    \frac{n-j-1}{2}+\alpha
    >0,
\end{equation}
so these counts also lie strictly below the threshold. Hence, both histories have a unique MAP action $0$ at each of the previous $n$ times. The two histories therefore share the same recent suffix of length $n+1$ and the same previous $n$ MAP actions, but have different current unique MAP actions. We now rule out global optimality in each recent-window class. First, consider $\mathcal W_B^{\mathrm{oa}}$, for which
\begin{equation}
    \nonumber
    n=n_B=\left\lfloor \frac{B}{2}\right\rfloor.
\end{equation}
Since $B\ge 3$ and $L=2^B-1$,
\begin{equation}
    \nonumber
    n_B+1<\frac{L}{2}+\alpha,
\end{equation}
so the construction above applies. Along both histories $r_{n_B}^-$ and $r_{n_B}^+$, the previous $n_B$ unique MAP actions are $0^{\,n_B}$. Hence, the current information variable
\begin{equation}
    \nonumber
    (G_t,R_t,R_{t-n_B:t-1},A_{t-n_B:t-1})
\end{equation}
is identical on the two histories. Any globally optimal policy in $\mathcal W_B^{\mathrm{oa}}$ therefore induces the same current action distribution on both histories. However, every globally optimal policy must choose the unique MAP action almost surely at every feasible history, and these unique MAP actions differ here. Thus, no policy in $\mathcal W_B^{\mathrm{oa}}$ is globally optimal. Now consider $\mathcal W_B^{\mathrm{obs}}$. If $B\ge 4$, take $n=B$. Then
\begin{equation}
    \nonumber
    B+1<\frac{L}{2}+\alpha,
\end{equation}
so the same construction applies. If $B=3$, use the explicit age-$7$ histories
\begin{equation}
    \nonumber
    r^- \triangleq (0,0,0,0,0,0,1),
    \qquad
    r^+ \triangleq (1,1,1,0,0,0,1),
\end{equation}
which have the same last $4$ status bits but different unique MAP actions because their total counts are $1$ and $4$. In either case, there are two feasible positive-probability histories with the same information variable
\begin{equation}
    \nonumber
    (G_t,R_t,R_{t-B:t-1})
\end{equation}
but different current unique MAP actions. Therefore, no policy in $\mathcal W_B^{\mathrm{obs}}$ is globally optimal. It remains to prove the strict inequality. For $\mathcal W_B^{\mathrm{obs}}$, define the finite information variable
\begin{equation}
    \nonumber
    X_t^{\mathrm{obs}} \triangleq (G_t,R_t,R_{t-B:t-1})
\end{equation}
taking values in
\begin{equation}
    \nonumber
    \mathcal{X}_B^{\mathrm{obs}}
     \triangleq 
    \{1,\ldots,L\}\times\{0,1\}\times\{0,1,\bot_R\}^{B},
\end{equation}
and let
\begin{equation}
    \nonumber
    \mathcal{P}_B^{\mathrm{obs}}
     \triangleq 
    \prod_{t\ge 0}\Delta(\{0,1\})^{\mathcal{X}_B^{\mathrm{obs}}}.
\end{equation}
For $\mathcal W_B^{\mathrm{oa}}$, define
\begin{equation}
    \nonumber
    X_t^{\mathrm{oa}}
     \triangleq 
    (G_t,R_t,R_{t-n_B:t-1},A_{t-n_B:t-1})
\end{equation}
and let $\mathcal{P}_B^{\mathrm{oa}}$ denote the analogous product of simplices. By Tychonoff's theorem, both policy spaces are compact. For each class $\mathcal W\in\{\mathcal W_B^{\mathrm{obs}},\mathcal W_B^{\mathrm{oa}}\}$, with compact policy space $\mathcal{P}_B^{\mathcal W}$, define the finite-horizon truncated value
\begin{equation}
    \nonumber
    V_e^{\mu}(T)=
    \E^\mu\!\left[\sum_{t=0}^{T}\gamma^{t}r(S_t,A_t)\right].
\end{equation}
The law of $(S_{0:T},Y_{0:T},A_{0:T})$ under $\mu$ depends only on the finitely many policy coordinates $(\mu_0,\ldots,\mu_T)$. Because all hidden and observed alphabets are finite, this truncated trajectory law is multilinear, hence continuous, in those finitely many simplex coordinates. Therefore $V_e^{\mu}(T)$ is continuous on $\mathcal{P}_B^{\mathcal W}$. Moreover, since rewards lie in $[0,1]$,
\begin{equation}
    \nonumber
    0\le V_e^{\mu}-V_e^{\mu}(T)
    \le \sum_{t=T+1}^\infty \gamma^{t}
    = \frac{\gamma^{T+1}}{1-\gamma}
\end{equation}
uniformly in $\mu$. Hence, $V_e^{\mu}(T)\to V_e^{\mu}$ uniformly, so $V_e^{\mu}$ is continuous on $\mathcal{P}_B^{\mathcal W}$ and therefore attains its maximum on each of the two classes. If, for either class $\mathcal W_B^{\mathrm{obs}}$ or $\mathcal W_B^{\mathrm{oa}}$, the attained maximum equaled $V_e^{\pi^{\mathrm{cnt}}}$, then a maximizing policy in that class would also be globally optimal, contradicting the exclusion just proved. Therefore
\begin{equation}
    \nonumber
    V_e^{\pi^{\mathrm{cnt}}}
    >
    \sup_{\pi\in\mathcal W_B^{\mathrm{obs}}} V_e^{\pi},
    \qquad
    V_e^{\pi^{\mathrm{cnt}}}
    >
    \sup_{\pi\in\mathcal W_B^{\mathrm{oa}}}  V_e^{\pi}.
\end{equation}
\end{proof}

\section{Proof of Theorem~\ref{thm:witness_lower_bound}}
\label{app:witness_lower_bound}
This section proves Theorem~\ref{thm:witness_lower_bound} and provides its extension over a horizon in Corollary~\ref{cor:summed_lower_bound}.

\begin{proof}[Proof of Theorem~\ref{thm:witness_lower_bound}.]
Fix $\kappa\in\Pi_B$. For each $0 \le t$, let
\begin{equation}
    \mathcal F_t\triangleq\sigma(Z_t,Y_t).
    \nonumber
\end{equation}
Define the Bellman-gap at $t$ by
\begin{equation}
    \nonumber
    \mathcal G_{e,t}^{\kappa}(B)
    \triangleq
    \E_e^\kappa\!\left[V_e^*(b_t) - \max_{a\in\cA}\E_e^\kappa\!\left[Q_e^*(b_t,a)\mid \mathcal F_t\right]\right].
\end{equation}
The controller observes $Y_0$ before the first action, so
\begin{equation}
    b_0(\cdot)=\Pp_e(S_0\in\cdot\mid Y_0),
    \qquad
    V_e^*=\E_e[V_e^*(b_0)].
    \nonumber
\end{equation}
The process induced by $\kappa$ is an admissible nonanticipative control process for the belief MDP and has the same discounted value as the original POMDP controller. For the belief MDP, write
\begin{equation}
    V_e^*-V_e^\kappa
    =
    \E_e^\kappa\!\left[\sum_{t=0}^\infty \gamma^t\bigl(V_e^*(b_t)-Q_e^*(b_t,A_t)\bigr)\right].
    \label{eq:canonical_converse_perf_diff}
\end{equation}
For each $0 \le t$, conditional on $\mathcal F_t$, the controller samples the action $A_t$ from the time-$t$ action kernel using a fresh exogenous seed. Hence
\begin{equation}
    A_t \perp b_t \mid \mathcal F_t.
    \nonumber
\end{equation}
Therefore,
\begin{align}
    \E_e^\kappa\!\left[Q_e^*(b_t,A_t)\mid \mathcal F_t\right]
    &=
    \sum_{a\in\cA}\pi_t(a\mid Z_t,Y_t)\,\E_e^\kappa\!\left[Q_e^*(b_t,a)\mid \mathcal F_t\right]
    \notag\\
    &\le
    \max_{a\in\cA}\E_e^\kappa\!\left[Q_e^*(b_t,a)\mid \mathcal F_t\right].
    \nonumber
\end{align}
Subtracting from $\E_e^\kappa[V_e^*(b_t)]$ yields
\begin{equation}
    \mathcal G_{e,t}^{\kappa}(B)
    \le
    \E_e^\kappa\!\left[V_e^*(b_t)-Q_e^*(b_t,A_t)\right].
\label{eq:canonical_converse_stage_bound}
\end{equation}
Since $V_e^*(b_t)=\max_{a\in\cA}Q_e^*(b_t,a)$, the summands in \eqref{eq:canonical_converse_perf_diff} are nonnegative. Tonelli's theorem and \eqref{eq:canonical_converse_stage_bound} give
\begin{equation}
    \gamma^t\,\mathcal G_{e,t}^{\kappa}(B)
    \le
    \sum_{t'=0}^\infty \gamma^{t'}\,\mathcal G_{e,t'}^{\kappa}(B)
    \le
    V_e^*-V_e^\kappa.
    \label{eq:canonical_converse_sum}
\end{equation}
If $\Pp_e^\kappa(E_t)=0$, then the left-hand side of \eqref{eq:witness_lower_bound} is $0$. Thus, the desired bound is immediate from \eqref{eq:canonical_converse_sum}. Assume that $\Pp_e^\kappa(E_t)>0$. Now fix the decision witness at time $t$. For each realization of $(Z_t,Y_t)$, choose a measurable selector
\begin{equation}
    A_t^\dagger \in
    \arg\max_{a\in\cA}\E_e^\kappa[Q_e^*(b_t,a)\mid \mathcal F_t].
    \nonumber
\end{equation}
Then
\begin{equation}
    \mathcal G_{e,t}^{\kappa}(B)
    =
    \E_e^\kappa\!\left[V_e^*(b_t)-Q_e^*(b_t,A_t^\dagger)\right].
    \nonumber
\end{equation}
Restricting to $E_t$ and using the suboptimality gap in Definition~\ref{def:decision_witness} gives
\begin{equation}
    \eta\,\Pp_e^\kappa(E_t)\,\Pp_e^\kappa\!\bigl(A_t^\dagger\neq a^*(\Theta_t,Y_t)\mid E_t\bigr)
    \le
    \mathcal G_{e,t}^{\kappa}(B).
    \label{eq:gap_to_adagger_error}
\end{equation}
For each fixed $y\in\cY_0$, the map $i\mapsto a^*(i,y)$ is injective. Thus, to choose the correct action on $E_t$, the label $\Theta_t$ should be correctly identified from $(Z_t,Y_t)$. Therefore
\begin{equation}
    \nonumber
    \inf_{\widehat\Theta}\Pp_e^\kappa\!\bigl(\widehat\Theta(Z_t,Y_t)\neq \Theta_t\mid E_t\bigr)
    \le
    \Pp_e^\kappa\!\bigl(A_t^\dagger\neq a^*(\Theta_t,Y_t)\mid E_t\bigr),
\end{equation}
where the infimum is over measurable decoders $\widehat\Theta:\supp(Z_t)\times\cY\to[N]$. Substituting into \eqref{eq:gap_to_adagger_error} gives
\begin{equation}
    \eta\,\Pp_e^\kappa(E_t)\,
    \inf_{\widehat\Theta}
    \Pp_e^\kappa\!\bigl(\widehat\Theta(Z_t,Y_t)\neq \Theta_t\mid E_t\bigr)
    \le
    \mathcal G_{e,t}^{\kappa}(B).
    \label{eq:witness_to_ambiguity_gap}
\end{equation}
It remains to lower-bound the conditional decoding error by Fano's inequality. Set
\begin{equation}
    \nonumber
    p_{\mathrm{err}}\triangleq\Pp_e^\kappa(\widehat\Theta\neq \Theta_t\mid E_t).
\end{equation}
Applying Fano's inequality under the conditional law given $E_t$ gives
\begin{equation}
    H_e^\kappa(\Theta_t\mid Z_t,Y_t,E_t)
    \le 1+p_{\mathrm{err}}\logtwo(N-1)
    \le 1+p_{\mathrm{err}}\logtwo N.
    \label{eq:fano_first}
\end{equation}
Write
\begin{equation}
    H_e^\kappa(\Theta_t\mid Z_t,Y_t,E_t)
    =
    H_e^\kappa(\Theta_t\mid Y_t,E_t)-I_e^\kappa(\Theta_t; Z_t\mid Y_t,E_t).
    \label{eq:fano_second}
\end{equation}
Combining \eqref{eq:fano_first} and \eqref{eq:fano_second} and rearranging yields
\begin{equation}
    \left(
    \frac{H_e^\kappa(\Theta_t\mid Y_t,E_t)-I_e^\kappa(\Theta_t; Z_t\mid Y_t,E_t)-1}{\logtwo N}\right)_+
    \le
    \Pp_e^\kappa(\widehat\Theta\neq \Theta_t\mid E_t).
    \label{eq:fano_substituted}
\end{equation}
Since $Z_t$ is itself an admissible $2^B$-ary summary of $\sH_t$, taking the infimum over all admissible summaries $P_{\Tilde Z_t\mid \sH_t}$, substituting \eqref{eq:fano_substituted} into \eqref{eq:witness_to_ambiguity_gap}, and then combining with \eqref{eq:canonical_converse_sum}, gives \eqref{eq:witness_lower_bound}. Taking the infimum over $\kappa\in\Pi_B$ then gives the claim for the lower bound on $D_e(B)$.
\end{proof}

\begin{corollary}[Summed decision witness lower bound]
\label{cor:summed_lower_bound}
Suppose that for each $t\ge 0$, there exists a decision witness such that, for every $\kappa\in\Pi_B$, the decision witness argument yields
\begin{equation}
    c_t^\kappa
    \le
    \E_e^\kappa\!\left[
    V_e^*(b_t)
    -
    \max_{a\in\cA}
    \E_e^\kappa\!\left[Q_e^*(b_t,a)\mid Z_t,Y_t\right]
    \right]
    \qquad\text{for some }0\le c_t^\kappa.
\end{equation}
Then
\begin{equation}
    \sum_{t=0}^\infty \gamma^t c_t^\kappa
    \le
    V_e^*-V_e^\kappa,
    \qquad
    \inf_{\kappa\in\Pi_B}\sum_{t=0}^\infty \gamma^t c_t^\kappa
    \le
    D_e(B).
\end{equation}
In particular, if the lower bound is uniform over $t$, i.e., $c\le c_t^\kappa$ for all $0\le t$ and all $\kappa\in\Pi_B$, then
\begin{equation}
    \frac{c}{1-\gamma}\le D_e(B).
\end{equation}
\end{corollary}

\begin{proof}
From the stagewise decomposition established in the proof of Theorem~\ref{thm:witness_lower_bound},
\begin{equation}
    \sum_{t=0}^\infty \gamma^t
    \E_e^\kappa\!\left[
    V_e^*(b_t)
    -
    \max_{a\in\cA}
    \E_e^\kappa\!\left[Q_e^*(b_t,a)\mid Z_t,Y_t\right]
    \right]
    \le
    V_e^*-V_e^\kappa.
    \nonumber
\end{equation}
Substituting the assumed lower bound and then taking the infimum over $\kappa\in\Pi_B$ gives the claim.
\end{proof}

\section{Proof of Proposition~\ref{prop:noisy_tracking_lower_bound}}
\label{app:tracking_example}
This section gives the full description of the noisy tracking problem and proves Proposition~\ref{prop:noisy_tracking_lower_bound}. In particular, we first derive the decision witness-based lower bound in Proposition~\ref{prop:noisy_tracking_lower_bound_app}. We then construct a finite-state controller in Definition~\ref{def:noisy_tracking_qp_fsc} and prove its gap upper bound in Proposition~\ref{prop:noisy_tracking_upper_bound}. Finally, Corollaries~\ref{cor:noisy_tracking_p_to_zero} and~\ref{cor:tracking_exact_frontier_p_zero_new} show that the lower and upper bounds match in the stated regimes.

\paragraph*{[Model.]} Fix integers $m\ge 1$ and $K\ge 2m+1$, and write
\begin{equation}
    \mathbb Z_K \triangleq \mathbb Z/K\mathbb Z.
    \nonumber
\end{equation}
Let
\begin{equation}
    \mathcal{D} \triangleq \{-m,-m+1,\ldots,m\}\subset\mathbb Z_K,
    \qquad
    L \triangleq |\mathcal{D}|=2m+1,
    \nonumber
\end{equation}
where $K\ge 2m+1$ makes the embedding of $\{-m,\ldots,m\}$ into $\mathbb Z_K$ injective. Fix parameters
\begin{equation}
    q\in(0,1),
    \qquad
    p\in[0,1),
    \nonumber
\end{equation}
and let $\bot\notin \mathbb Z_K\cup \mathcal{D}$ be an erasure symbol. Define the alphabets
\begin{equation}
    \cS\triangleq\{\mathrm b,\mathrm o\}\times\{0,1\}\times \mathbb Z_K\times \mathcal{D},
    \quad
    \cY \triangleq \bigl(\{\mathrm b\}\times (\mathbb Z_K\cup\{\bot\})\bigr)\cup \bigl(\{\mathrm o\}\times (\mathcal{D} \cup\{\bot\})\bigr),
    \quad
    \cA \triangleq \mathbb Z_K.
    \nonumber
\end{equation}
The hidden state is
\begin{equation}
    \nonumber
    S_t=(C_t,R_t,X_t,D_t)\in\cS,
\end{equation}
where $X_t$ is the current position, $D_t$ is the current increment, $C_t$ indicates whether the current observation is a beacon ($\mathrm b$) or odometry ($\mathrm o$), and $R_t$ indicates whether the current reading is reliable. Initialize with
\begin{equation}
    X_0\sim\mathrm{Unif}(\mathbb Z_K),
    \qquad
    D_0=0,
    \qquad
    C_0=\mathrm b,
    \qquad
    R_0=1.
    \nonumber
\end{equation}
For every $0 \le t$, conditional on the current state and action $a$, draw independently
\begin{equation}
    \nonumber
    D_{t+1}\sim\mathrm{Unif}(\mathcal{D}),
\end{equation}
\begin{equation}
    \nonumber
    C_{t+1}=
    \begin{cases}
        \mathrm b,& \text{with probability }q,\\
        \mathrm o,& \text{with probability }1-q,
    \end{cases}
    \qquad
    R_{t+1}=
    \begin{cases}
        1,& \text{with probability }1-p,\\
        0,& \text{with probability }p,
    \end{cases}
\end{equation}
and set
\begin{equation}
    \nonumber
    X_{t+1}=X_t+D_{t+1}\pmod K.
\end{equation}
Thus, the state transition kernel is independent of the control action. At time $t$, the controller observes $Y_t\in\cY$ with deterministic observation kernel
\begin{equation}
    Y_t=
    \begin{cases}
        (\mathrm b,X_t),& C_t=\mathrm b,\ R_t=1,\\
        (\mathrm b,\bot),& C_t=\mathrm b,\ R_t=0,\\
        (\mathrm o,D_t),& C_t=\mathrm o,\ R_t=1,\\
        (\mathrm o,\bot),& C_t=\mathrm o,\ R_t=0.
    \end{cases}
    \nonumber
\end{equation}
Hence, the controller receives either a clean beacon, a clean odometry reading, or an erasure with probability $p$ in both modes. The one-step reward is exact localization:
\begin{equation}
    \nonumber
    r(S_t,a)=\ind\{a=X_t\}.
\end{equation}

\begin{proposition}[Tracking example lower bound]
\label{prop:noisy_tracking_lower_bound_app}
Consider the noisy tracking model above and let $e=e(K,m,p,q)$ denote the corresponding environment. Then
\begin{equation}
    \sum_{t\ge 1}\gamma^t\,\alpha_t\left(1-\frac{2^B}{K}\right)_+
    \le
    D_e(B),
    \quad \textrm{ where } \
    \alpha_t \triangleq \rho^t+q(1-p)\sum_{t'=1}^{t-1}\rho^{t'},
    \quad
    \rho \triangleq  (1-q)(1-p).
    \label{eq:noisy_tracking_frontier_app}
\end{equation}
\end{proposition}

\begin{proof}
Fix $\kappa\in\Pi_B$ and $1\le t$. We instantiate the decision witness from Definition~\ref{def:decision_witness} as follows. Let $\iota:\mathbb Z_K\to[K]$ be a fixed bijection. Define
\begin{equation}
    \psi(b) \triangleq 
    \begin{cases}
        \iota(x), & b\bigl(\{s\in\cS: x(s)=x\}\bigr)=1 
        \text{ for some } x\in\mathbb Z_K,\\
        0, & \text{otherwise.}
    \end{cases}
    \nonumber
\end{equation}
Set
\begin{equation}
    \Theta_t\triangleq\psi(b_t),
    \qquad
    \cY_{\mathrm{odo}} \triangleq \{(\mathrm o,d):d\in \mathcal{D}\},
    \qquad
    \cY_{\mathrm{bea}} \triangleq \{(\mathrm b,x):x\in\mathbb Z_K\}.
    \nonumber
\end{equation}
Define the witness event
\begin{equation}
    E_t\triangleq \{\Theta_t\in[K],\ Y_t\in\cY_{\mathrm{odo}}\},
    \nonumber
\end{equation}
and action map
\begin{equation}
    a^*:[K]\times \cY_{\mathrm{odo}}\to \cA,
    \qquad
    a^*(i,(\mathrm o,d))\triangleq\iota^{-1}(i).
    \nonumber
\end{equation}
For each fixed $y\in\cY_{\mathrm{odo}}$, the map $i\mapsto a^*(i,y)$ is injective. On $E_t$, the exact belief $b_t$ pins the position to a unique $x=X_t$, so $Q_e^*(b_t,a)=\ind\{a=X_t\}+\gamma\E_e[V_e^*(b_{t+1})\mid b_t]$ for every $a\in\cA$, giving
\begin{equation}
    V_e^*(b_t)-Q_e^*(b_t,a)=\ind\{a\neq a^*(\Theta_t,Y_t)\} \qquad\forall a\in\cA,\quad\text{on }E_t.
    \label{eq:tracking_gap} 
\end{equation}
This verifies the witness gap condition with $\eta=1$. We decompose the witness event $E_t$ into sub-events. First, for each $d\in \mathcal{D}$, let
\begin{equation}
    \nonumber
    F_{0,t}^d \triangleq 
    \begin{cases}
        \{Y_1=(\mathrm o,d)\}, & t=1,\\
        \{Y_1\in\cY_{\mathrm{odo}},\ldots,Y_{t-1}\in\cY_{\mathrm{odo}},\ Y_t=(\mathrm o,d)\}, & t\ge 2.
    \end{cases}
    \nonumber
\end{equation}
Then, for $t'\in\{1,\ldots,t-1\}$,
\begin{equation}
    F_{t',t}^d \triangleq \{Y_{t'}\in\cY_{\mathrm{bea}},\ Y_{t'+1}\in\cY_{\mathrm{odo}},\ldots,Y_{t-1}\in\cY_{\mathrm{odo}},\ Y_t=(\mathrm o,d)\}.
    \nonumber
\end{equation}
We have
\begin{equation}
    E_t = \bigcup_{d\in \mathcal{D}}E_t^d,
    \qquad
    E_t^d \triangleq \bigcup_{t'=0}^{t-1}F_{t',t}^d.
    \nonumber
\end{equation}
We use two immediate consequences of this decomposition. First, on $F_{0,t}^d$, the initial beacon $Y_0=(\mathrm b,X_0)$ reveals $X_0$ exactly, and $Y_1,\ldots,Y_t$ are all clean odometry observations, so
\begin{equation}
    X_{t''}=X_{t''-1}+D_{t''}\pmod K,
    \qquad t''=1,\ldots,t,
    \nonumber
\end{equation}
determine $X_t$ exactly from $Y_{0:t}$. Similarly, on $F_{t',t}^d$ with $t'\in\{1,\ldots,t-1\}$, the clean beacon $Y_{t'}=(\mathrm b,X_{t'})$ reveals $X_{t'}$ exactly, and the subsequent clean odometry observations determine $X_t$ through
\begin{equation}
    \nonumber
    X_{t''}=X_{t''-1}+D_{t''}\pmod K,
    \qquad t''=t'+1,\ldots,t.
\end{equation}
Hence, on $E_t^d$ the exact belief $b_t$ determines the current position exactly. There exists a unique $x=X_t$ such that
\begin{equation}
    b_t\bigl(\{s\in\cS:x(s)=x\}\bigr)=1.
    \label{eq:noisy_tracking_refined_position_mass}
\end{equation}
Second, because the events $F_{t',t}^d$ are disjoint and depend only on the mode/reliability process and the current odometry symbol, we have
\begin{equation}
    \Pp_e^\kappa(F_{0,t}^d)=\frac{\rho^t}{L},
    \qquad
    \Pp_e^\kappa(F_{t',t}^d)=\frac{q(1-p)\rho^{t-t'}}{L},
    \quad t'=1,\ldots,t-1.\nonumber
\end{equation}
Summing over $t'$ and $d\in \mathcal{D}$ yields
\begin{equation}
    \Pp_e^\kappa(E_t)=\alpha_t.
    \nonumber
\end{equation}
These probabilities are independent of $\kappa$. We now show that
\begin{equation}
    X_t\mid E_t^d\sim \mathrm{Unif}(\mathbb Z_K).
    \label{eq:noisy_tracking_refined_uniformity}
\end{equation}
Fix $c\in\mathbb Z_K$. Let $T_c$ add $c$ modulo $K$ only to the position coordinates of state trajectories, and let $\Gamma_c$ add $c$ modulo $K$ only to beacon observations, leaving odometry observations unchanged. Since $X_0$ is uniform on $\mathbb Z_K$, the state recursion is additive, and the mode, reliability, and increment processes are independent of the absolute position, the joint law of $(S_{0:t},Y_{0:t})$ is equivariant under $(T_c,\Gamma_c)$. Each event $F_{t',t}^d$ is invariant under $\Gamma_c$ because it constrains only the beacon/odometry type, reliability, and current odometry symbol $d$, not the absolute position. Hence, $E_t^d$ is invariant. For any $x,x'\in\mathbb Z_K$, choose $c=x'-x$. Then $(T_c,\Gamma_c)$ maps $\{X_t=x\}\cap E_t^d$ bijectively onto $\{X_t=x'\}\cap E_t^d$ and preserves probability. Therefore
\begin{equation}
    \Pp_e^\kappa(X_t=x,\,E_t^d)=\Pp_e^\kappa(X_t=x',\,E_t^d)
    \qquad\forall x,x'\in\mathbb Z_K.
    \nonumber
\end{equation}
Since the $K$ masses sum to $\Pp_e^\kappa(E_t^d)$, each equals $\Pp_e^\kappa(E_t^d)/K$, proving \eqref{eq:noisy_tracking_refined_uniformity}. Choose a deterministic measurable selector
\begin{equation}
    A_t^\dagger\in\arg\max_{a\in\cA}\E_e^\kappa\!\left[Q_e^*(b_t,a)\mid Z_t,Y_t\right].
    \nonumber
\end{equation}
Define 
\begin{equation}
    \mathcal G_{e,t}^\kappa(B)\triangleq \E_e^\kappa[V_e^*(b_t)-Q_e^*(b_t,A_t^\dagger)].
    \nonumber
\end{equation}
On $E_t^d$, the observation $Y_t=(\mathrm{o},d)$ is fixed, so 
$A_t^\dagger$ is a deterministic function of $Z_t$ alone. Since $Z_t$ has at most $2^B$ possible values, any decoder based only on $Z_t$ can correctly identify at most $2^B$ of the $K$ equally likely positions. Therefore,
\begin{equation}
\Pp_e^\kappa(A_t^\dagger=X_t\mid E_t^d)
\le
\frac{2^B}{K}.
\label{eq:success_prob}
\end{equation}
Substituting~\eqref{eq:success_prob} into~\eqref{eq:tracking_gap} yields
\begin{equation}
    \left(1-\frac{2^B}{K}\right)_+
    \le 
    \Pp_e^\kappa(A_t^\dagger\neq X_t\mid E_t^d)
    =
    \E_e^\kappa\!\left[V_e^*(b_t)-Q_e^*(b_t,A_t^\dagger)\mid E_t^d\right].
    \nonumber
\end{equation}
Multiplying by $\Pp_e^\kappa(E_t^d)=\alpha_t/L$ and summing over $d\in \mathcal{D}$ yields
\begin{equation}
    \nonumber
    \alpha_t\left(1-\frac{2^B}{K}\right)_+
    =
    \sum_{d\in \mathcal{D}}\Pp_e^\kappa(E_t^d)\left(1-\frac{2^B}{K}\right)_+
    \le
    \mathcal G_{e,t}^{\kappa}(B).
\end{equation}
Since this holds for every $\kappa\in\Pi_B$ and every $t\ge 1$,
\begin{equation}
    \nonumber
    \sum_{t \ge 1}\gamma^t\,\alpha_t\left(1-\frac{2^B}{K}\right)_+
    \le
    \sum_{t\ge 0}\gamma^t\,\mathcal G_{e,t}^{\kappa}(B)
    \le
    V_e^*-V_e^\kappa.
\end{equation}
Taking the infimum over $\kappa$ yields \eqref{eq:noisy_tracking_frontier_app}.
\end{proof}

\begin{definition}[Quotient-propagation FSC]
\label{def:noisy_tracking_qp_fsc}
Assume from now on that
\begin{equation}
    2^B\mid K.
\label{eq:noisy_tracking_divisibility}
\end{equation}
Let $G\le \mathbb Z_K$ be the subgroup of size $K/2^B$, and let
\begin{equation}
    q_G:\mathbb Z_K\to\mathbb Z_K/G
    \nonumber
\end{equation}
be the quotient map. The controller stores only the coset of the position $X$. Since every coset has size $K/2^B$, the quotient $\mathbb Z_K/G$ has exactly $2^B$ elements and therefore fits the $B$-bit memory budget. We define the quotient addition by
\begin{equation}
    (x+G)+(d+G) \triangleq (x+d)+G,
    \nonumber
\end{equation}
so the quotient map respects motion updates:
\begin{equation}
    q_G(x+d)=q_G(x)+q_G(d),\qquad x\in\mathbb Z_K,\ d\in \mathcal{D}.
    \label{eq:quotient_sum}
\end{equation}
Fix a representative selector
\begin{equation}
    \nonumber
    \mathrm{rep}_G:\mathbb Z_K/G\to\mathbb Z_K,
    \qquad \mathrm{rep}_G(z)\in z\quad\forall z\in\mathbb Z_K/G,
\end{equation}
and define the deterministic quotient-update map
\begin{equation}
    T_G(z,y) \triangleq 
    \nonumber
    \begin{cases}
    q_G(x), & y=(\mathrm b,x),\\
    z+q_G(d), & y=(\mathrm o,d),\\
    z, & y\in\{(\mathrm b,\bot),(\mathrm o,\bot)\}.
    \end{cases}
\end{equation}
The quotient-propagation FSC $\kappa_{B,p}^{\mathrm{qp}}\in\Pi_B$ is an FSC with memory alphabet
\begin{equation}
    \cZ \triangleq \mathbb Z_K/G,
    \nonumber
\end{equation}
initialization kernel given by any fixed point mass on $\cZ$, action rule
\begin{equation}
    A_t=
    \begin{cases}
        x, & Y_t=(\mathrm b,x),\\
        \mathrm{rep}_G\bigl(Z_t+q_G(d)\bigr), & Y_t=(\mathrm o,d),\\
        \mathrm{rep}_G(Z_t), & Y_t\in\{(\mathrm b,\bot),(\mathrm o,\bot)\},
    \end{cases}
    \label{eq:noisy_tracking_qp_action}
\end{equation}
and deterministic transition kernel
\begin{equation}
    \varphi_t(z'\mid z,y,a)=\ind \{z'=T_G(z,y)\}.
    \label{eq:noisy_tracking_qp_kernel}
\end{equation}
Equivalently, the controller stores a quotient-valued position summary: a clean beacon resets it to the observed quotient, clean odometry propagates it by $q_G(d)$, and an erasure leaves it unchanged.
\end{definition}

\begin{proposition}[Quotient-propagation FSC upper bound]
\label{prop:noisy_tracking_upper_bound}
Assume \eqref{eq:noisy_tracking_divisibility}. The quotient-propagation FSC $\kappa_{B,p}^{\mathrm{qp}}$ provides a construction satisfying
\begin{equation}
    D_e(B)
    \le
    \sum_{t\ge 1}\gamma^t\left[1-q(1-p)-\alpha_t\frac{2^B}{K}\right].
    \label{eq:noisy_tracking_upper_regret}
\end{equation}
\end{proposition}

\begin{proof}
At time $0$, the clean initial beacon $Y_0=(\mathrm b,X_0)$ makes the controller choose $A_0=X_0$ and receive reward $1$. Fix $t\ge 1$. Define $\cY_{\mathrm{odo}} \triangleq \{(\mathrm o,d):d\in \mathcal{D}\}$ and $\cY_{\mathrm{bea}} \triangleq \{(\mathrm b,x):x\in\mathbb Z_K\}$. For each $d\in \mathcal{D}$, define
\begin{equation}
    F_{0,t}^d \triangleq 
    \nonumber
    \begin{cases}
        \{Y_1=(\mathrm o,d)\}, & t=1,\\
        \{Y_1\in\cY_{\mathrm{odo}},\ldots,Y_{t-1}\in\cY_{\mathrm{odo}},\ Y_t=(\mathrm o,d)\}, & t\ge 2,
        \end{cases}
\end{equation}
and, for $t'\in\{1,\ldots,t-1\}$,
\begin{equation}
    \nonumber
    F_{t',t}^d \triangleq \{Y_{t'}\in\cY_{\mathrm{bea}},\ Y_{t'+1}\in\cY_{\mathrm{odo}},\ldots,Y_{t-1}\in\cY_{\mathrm{odo}},\ Y_t=(\mathrm o,d)\}.
\end{equation}
Set
\begin{equation}
    E_t^d \triangleq \bigcup_{t'=0}^{t-1}F_{t',t}^d,
    \qquad
    E_t \triangleq \bigcup_{d\in \mathcal{D}}E_t^d.
    \nonumber
\end{equation}
Here, these match the decision witness slices used in the lower-bound proof. Fix $d\in \mathcal{D}$ and a sample path in $E_t^d$. By definition of $E_t^d$, there exists a unique $t'\in\{0,1,\ldots,t-1\}$ such that either $t'=0$ and $Y_1,\ldots,Y_t$ are all clean odometry observations, or $t'\ge 1$, $Y_{t'}$ is a clean beacon, and $Y_{t'+1},\ldots,Y_t$ are clean odometry observations. If $t'=0$, then the time-$0$ beacon is clean, so the update at time $0$ yields
\begin{equation}
    \nonumber
    Z_1=T_G(Z_0,Y_0)=q_G(X_0).
\end{equation}
We now prove by induction that
\begin{equation}
    \nonumber
    Z_{t''}=q_G(X_{t''-1}),\qquad t''=1,2,\ldots,t.
\end{equation}
The base case $t''=1$ appears above. For the induction step, suppose $1\le t''<t$ and $Z_{t''}=q_G(X_{t''-1})$. On $E_t^d$, the observation $Y_{t''}$ is a clean odometry symbol $(\mathrm o,d_{t''})$ for some $d_{t''}\in \mathcal{D}$. By \eqref{eq:quotient_sum}, the FSC update gives
\begin{equation}
    \nonumber
    Z_{t''+1}=T_G(Z_{t''},Y_{t''})=T_G(q_G(X_{t''-1}),(\mathrm o,d_{t''}))=q_G(X_{t''-1}+d_{t''})=q_G(X_{t''}).
\end{equation}
Thus, the claim holds for all $t''\le t$, and in particular $Z_t=q_G(X_{t-1})$. If $t'\ge 1$, then the clean beacon at time $t'$ identifies $X_{t'}$ exactly, and after processing that beacon the update yields
\begin{equation}
    \nonumber
    Z_{t'+1}=T_G(Z_{t'},Y_{t'})=q_G(X_{t'}).
\end{equation}
Starting from this base case, the same induction as above shows that $Z_{t''}=q_G(X_{t''-1})$ for every $t''=t'+1,\ldots,t$. Thus, again $Z_t=q_G(X_{t-1})$. On $E_t^d$, the identity $Z_t=q_G(X_{t-1})$ and the clean odometry observation $Y_t=(\mathrm o,d)$ imply
\begin{equation}
    \nonumber
    A_t=\mathrm{rep}_G\bigl(Z_t+q_G(d)\bigr)=\mathrm{rep}_G\bigl(q_G(X_{t-1})+q_G(d)\bigr)=\mathrm{rep}_G\bigl(q_G(X_t)\bigr).
\end{equation}
To verify the conditional uniformity, fix $d\in \mathcal{D}$ and $c\in\mathbb Z_K$. Let $T_c$ shift only the position coordinate of the state trajectory by $c$ modulo $K$, and let $\Gamma_c$ shift only beacon observations by $c$, leaving odometry observations unchanged. Because the state recursion is additive and the mode, reliability, and increment processes are independent of the absolute position, the joint law of $(S_{0:t},Y_{0:t})$ is equivariant under $(T_c,\Gamma_c)$. The event $E_t^d$ is invariant under $\Gamma_c$ because it constrains only the beacon/odometry pattern and the current clean odometry symbol $d$, not the absolute position. Therefore $(T_c,\Gamma_c)$ maps $\{X_t=x\}\cap E_t^d$ bijectively onto $\{X_t=x+c\}\cap E_t^d$ while preserving probability, so the masses $\Pp(X_t=x,E_t^d)$ are all equal in $x$. Since they sum to $\Pp(E_t^d)$, we obtain
\begin{equation}
    \nonumber
    X_t\mid E_t^d\sim \mathrm{Unif}(\mathbb Z_K).
\end{equation}
Since every coset of $G$ has the same cardinality $|G|=K/2^B$, conditioning on the event $\{q_G(X_t)=z\}$ makes $X_t$ uniform on that coset. Therefore, the probability that the fixed representative equals the true position is
\begin{equation}
    \nonumber
    \Pp\!\left(A_t=X_t\mid E_t^d,\ q_G(X_t)=z,\ \kappa_{B,p}^{\mathrm{qp}}\right)
    =\frac{1}{|G|}=\frac{2^B}{K}
    \qquad \forall z\in \mathbb Z_K/G,
\end{equation}
and averaging over the coset label gives
\begin{equation}
    \nonumber
    \Pp\!\left(A_t=X_t\mid E_t^d,\ \kappa_{B,p}^{\mathrm{qp}}\right)=\frac{2^B}{K}.
\end{equation}
On the clean-beacon event $Y_t\in\cY_{\mathrm{bea}}$, the first case of the action rule \eqref{eq:noisy_tracking_qp_action} gives
\begin{equation}
    \nonumber
    \Pp\!\left(A_t=X_t\mid Y_t\in\cY_{\mathrm{bea}},\ \kappa_{B,p}^{\mathrm{qp}}\right)=1.
\end{equation}
The events $\{Y_t\in\cY_{\mathrm{bea}}\}$ and $E_t$ are disjoint, because the former requires the current observation to be a clean beacon while the latter requires it to be a clean odometry symbol. Moreover, the slice events $E_t^d$ are pairwise disjoint because the current clean odometry symbol $d$ is fixed on each slice. On the clean beacon event $Y_t\in\cY_{\mathrm{bea}}$, the controller is exact; this event has probability $q(1-p)$. On the odometry witness $E_t=\bigcup_{d\in \mathcal{D}}E_t^d$, the slice-wise success probability is the same for every $d\in \mathcal{D}$, so
\begin{align}
    \Pp\!\left(A_t=X_t\mid E_t,\ \kappa_{B,p}^{\mathrm{qp}}\right)
    &=
    \sum_{d\in \mathcal{D}}\Pp(E_t^d\mid E_t)\,\Pp\!\left(A_t=X_t\mid E_t^d,\ \kappa_{B,p}^{\mathrm{qp}}\right)
    \nonumber\\
    &=
    \sum_{d\in \mathcal{D}}\Pp(E_t^d\mid E_t)\,\frac{2^B}{K}
    =\frac{2^B}{K}.
    \nonumber
\end{align}
By the same event-probability calculation as in the converse,
\begin{equation}
    \nonumber
    \Pp(E_t)=\alpha_t.
\end{equation}
Since the reward is nonnegative on the complement of these two events, the expected stage-$t$ reward is at least
\begin{equation}
    \nonumber
    q(1-p)+\alpha_t\frac{2^B}{K}.
\end{equation}
Summing over $t\ge 1$ together with the exact stage-$0$ reward gives
\begin{equation}
    \nonumber
    1+\sum_{t\ge 1}\gamma^t\left[q(1-p)+\alpha_t\frac{2^B}{K}\right]
    \le
    V_e^{\kappa_{B,p}^{\mathrm{qp}}}.
\end{equation}
Finally,
\begin{equation}
    \nonumber
    V_e^*\le 1+\sum_{t\ge 1}\gamma^t,
\end{equation}
because each stage reward is at most $1$. Subtracting the lower bound on $V_e^{\kappa_{B,p}^{\mathrm{qp}}}$ from this trivial upper bound on $V_e^*$ yields \eqref{eq:noisy_tracking_upper_regret}.
\end{proof}

\begin{corollary}[Sharpness as $p\to 0$]
\label{cor:noisy_tracking_p_to_zero}
Assume \eqref{eq:noisy_tracking_divisibility}. Then, for every fixed budget $B$, the converse in Proposition~\ref{prop:noisy_tracking_lower_bound_app} and the quotient-propagation upper bound in Proposition~\ref{prop:noisy_tracking_upper_bound} converge to the same $p=0$ frontier:
\begin{equation}
    \lim_{p\to 0}
    \sum_{t\ge 1}\gamma^t\,\alpha_t\left(1-\frac{2^B}{K}\right)
    =
    \lim_{p\to 0}
    \sum_{t\ge 1}\gamma^t\left[1-q(1-p)-\alpha_t\frac{2^B}{K}\right]
    =
    \frac{\gamma(1-q)}{1-\gamma}\left(1-\frac{2^B}{K}\right).
\label{eq:noisy_tracking_p_to_zero_limits}
\end{equation}
\end{corollary}

\begin{proof}
Let $\rho=(1-q)(1-p)$. Then $\rho\to 1-q$ and $q(1-p)\to q$ as $p\to 0$. From the definition
\begin{equation}
    \nonumber
    \alpha_t=\rho^t+q(1-p)\sum_{t'=1}^{t-1}\rho^{t'}
\end{equation}
in Proposition~\ref{prop:noisy_tracking_lower_bound_app}, for each fixed $t$,
\begin{equation}
    \nonumber
    \alpha_t\longrightarrow (1-q)^t+q\sum_{t'=1}^{t-1}(1-q)^{t'}=1-q.
\end{equation}
Since $0\le \alpha_t\le 1$ for all $t$ and $p$,
\begin{equation}
    \nonumber
    \sum_{t\ge 1}\gamma^t\alpha_t\longrightarrow (1-q)\sum_{t\ge 1}\gamma^t = \frac{\gamma(1-q)}{1-\gamma}.
\end{equation}
Applying the same argument to the upper-bound summand $1-q(1-p)-\alpha_t 2^B/K$ gives \eqref{eq:noisy_tracking_p_to_zero_limits}.
\end{proof}

\begin{corollary}[Tightness at $p=0$]
\label{cor:tracking_exact_frontier_p_zero_new}
Assume $p=0$ and \eqref{eq:noisy_tracking_divisibility}. Then
\begin{equation}
    D_e(B)=\frac{\gamma(1-q)}{1-\gamma}\left(1-\frac{2^B}{K}\right),
    \label{eq:tracking_exact_frontier_regret_p_zero_new}
\end{equation}
which is equivalent to
\begin{equation}
    V_e^*(B)=1+\sum_{t\ge 1}\gamma^t\left[q+(1-q)\frac{2^B}{K}\right].
    \label{eq:tracking_exact_frontier_value_p_zero_new}
\end{equation}
\end{corollary}

\begin{proof}
Setting $p=0$ in Proposition~\ref{prop:noisy_tracking_lower_bound_app} gives the lower bound
\begin{equation}
    \nonumber
    D_e(B)\ge \frac{\gamma(1-q)}{1-\gamma}\left(1-\frac{2^B}{K}\right),
\end{equation}
because then $\alpha_t=1-q$ for every $t\ge 1$. Setting $p=0$ in Proposition~\ref{prop:noisy_tracking_upper_bound} gives the matching upper bound
\begin{equation}
    \nonumber
    D_e(B)\le \frac{\gamma(1-q)}{1-\gamma}\left(1-\frac{2^B}{K}\right).
\end{equation}
Hence
\begin{equation}
\nonumber
D_e(B)= \frac{\gamma(1-q)}{1-\gamma}\left(1-\frac{2^B}{K}\right),
\end{equation}
which is \eqref{eq:tracking_exact_frontier_regret_p_zero_new}. For \eqref{eq:tracking_exact_frontier_value_p_zero_new}, use
\begin{equation}
    \nonumber
    V_e^*(B)=V_e^*-D_e(B).
\end{equation}
When $p=0$, the stage-$0$ reward is exactly $1$, and from stage $t\ge 1$ onward the fully informed controller is correct with probability $q+(1-q)=1$. Thus
\begin{equation}
    \nonumber
    V_e^*=1+\sum_{t\ge 1}\gamma^t.
\end{equation}
Substituting the expression for $D_e(B)$ gives
\begin{align}
    \nonumber
    V_e^*(B)
    &=
    \nonumber
    1+\sum_{t\ge1}\gamma^t
    -\frac{\gamma(1-q)}{1-\gamma}\left(1-\frac{2^B}{K}\right) \\
    &=
    \nonumber
    1+\sum_{t\ge1}\gamma^t
    -\sum_{t\ge1}\gamma^t(1-q)\left(1-\frac{2^B}{K}\right) \\
    &=
    \nonumber
    1+\sum_{t\ge1}\gamma^t
    \left[
    1-(1-q)\left(1-\frac{2^B}{K}\right)
    \right] \\
    &=
    1+\sum_{t\ge1}\gamma^t
    \left[
    q+(1-q)\frac{2^B}{K}
    \right].
    \nonumber
\end{align}
This is exactly \eqref{eq:tracking_exact_frontier_value_p_zero_new}.
\end{proof}
Therefore, the exact-localization decision witnesses prove the lower bound \eqref{eq:noisy_tracking_frontier}. When $p=0$ and $2^B\mid K$, the quotient-propagation FSC $\kappa_B^{\mathrm{qp}}\triangleq\kappa_{B,0}^{\mathrm{qp}}$ attains the same gap, making the lower bound tight. This proves Proposition~\ref{prop:noisy_tracking_lower_bound}. 

\section{Decision Witness Example: Hidden Source}
\label{app:witness_example_ii}
This section gives another example of a decision witness construction. This time, it is for a finite-state POMDP with a hidden source.
\paragraph*{[Model.]}
Fix integers $K\ge 2$ and $n\ge 2$, a jump probability $\alpha\in(0,1)$, and parameters
\begin{equation}
    \nonumber
    0<p_1<\cdots<p_K<1.
\end{equation}
Define the alphabets
\begin{equation}
    \nonumber
    \cS_{\mathrm{aoi}}  \triangleq  [K]\times [n],
    \qquad
    \cY_{\mathrm{aoi}}  \triangleq  [n]\times \{0,1\},
    \qquad
    \cA_{\mathrm{aoi}}   \triangleq  [K].
\end{equation}
The hidden state is
\begin{equation}
    \nonumber
    S_t  \triangleq  (J_t,\tau_t)\in \cS_{\mathrm{aoi}},
\end{equation}
where $J_t\in[K]$ is the active hidden source label and $\tau_t\in[n]$ is the observed age of the current source episode with a forced refresh after age $n$. Initialize at time $t=0$ with
\begin{equation}
    \nonumber
    J_0 \sim \mathrm{Unif}([K]),
    \qquad
    \tau_0 = 1.
\end{equation}
For every $t\ge 0$, conditional on the current state and action $a$, the next state is drawn as follows:
\begin{equation}
    \nonumber
    (J_{t+1},\tau_{t+1})=
    \begin{cases}
        (J_t,\tau_t+1), & \tau_t<n\ \text{ and no jump with prob. } 1-\alpha,\\
        (\widetilde J_{t+1},1),\quad \widetilde J_{t+1}\sim \mathrm{Unif}([K]\setminus\{J_t\}),
        & \tau_t<n\ \text{and a jump with prob. }\alpha,\\
        (\widetilde J_{t+1},1),\quad \widetilde J_{t+1}\sim \mathrm{Unif}([K]\setminus\{J_t\}),
        & \tau_t=n,
    \end{cases}
    \nonumber
\end{equation}
independent of the control action. Thus, the active source label evolves independently of the action until the age reaches $n$, after which a refresh is forced on the next transition; $\tau_t$ records the age of the current source episode. At time $t$, the controller observes
\begin{equation}
    Y_t  \triangleq  (\tau_t,O_t)\in \cY_{\mathrm{aoi}},
    \nonumber
\end{equation}
where $O_t\in\{0,1\}$ is a noisy source bit satisfying
\begin{equation}
    \nonumber
    \Pp(O_t=1\mid J_t=i)=p_i,
    \qquad i\in[K].
\end{equation}
The controller chooses an action
\begin{equation}
    \nonumber
    A_t\in\cA_{\mathrm{aoi}}
\end{equation}
and receives the one-step reward
\begin{equation}
    r(S_t,A_t) \triangleq v(\tau_t)\,\ind\{A_t=J_t\},
    \nonumber
\end{equation}
where $v:[n]\to(0,1]$ is any positive function. Because both the state transition and the observation kernels are action-independent, the marginal law of $(S_{0:t},Y_{0:t})$ is the same under every controller $\kappa\in\Pi_B$. In particular, the exact belief process $(b_t)$ is a deterministic measurable function of the observation history.

\begin{proposition}[Decision witness lower bound for hidden-source]
\label{prop:aoi_markov_gap}
Consider the hidden-source model above. Let $e=e(K,n,\alpha,p_1,\ldots,p_K,v)$ denote the corresponding environment. Let $\delta>0$ satisfy
\begin{equation}
    4\delta < \min_{i\neq j}|p_i-p_j|,
    \qquad
    [p_i-\delta,p_i+\delta]\subset(0,1)\ \ \forall i\in[K],
    \label{eq:aoi_markov_delta}
\end{equation}
and
\begin{equation}
    r\logtwo p_i+(1-r)\logtwo(1-p_i)
    >
    r\logtwo p_j+(1-r)\logtwo(1-p_j)
    \quad
    \forall r\in[p_i-\delta,p_i+\delta],\ \forall i\neq j.
\end{equation}
Set
\begin{equation}
    \xi  \triangleq 
    \min_{i\neq j}
    \inf_{r\in[p_i-\delta,p_i+\delta]}
    \Bigl(
    r\logtwo p_i+(1-r)\logtwo(1-p_i)
    -
    r\logtwo p_j-(1-r)\logtwo(1-p_j)
    \Bigr),
    \label{eq:aoi_markov_xi}
\end{equation}
and define $a_n  \triangleq  (K-1)2^{-n\xi}$. Assume throughout that $a_n<1$. Then
\begin{equation}
    \gamma^{n-1}(1-\alpha)^{n-1}
    \,v(n)\frac{1-a_n}{1+a_n}
    \left(1-2e^{-2n\delta^2}-\frac{B+2}{\logtwo K}\right)_+
    \le
    D_e(B).
\label{eq:aoi_markov_frontier}
\end{equation}
\end{proposition}

\begin{proof}
Define
\begin{equation}
    \nonumber
    f_p(r) \triangleq r\logtwo p+(1-r)\logtwo(1-p),
    \qquad
    g_{ij}(r) \triangleq f_{p_i}(r)-f_{p_j}(r),
    \qquad r\in[0,1],\ p\in(0,1).
\end{equation}
By continuity of each $g_{ij}$ and the fact that
\begin{equation}
    \nonumber
    g_{ij}(p_i)=D_2(\mathrm{Bern}(p_i)\|\mathrm{Bern}(p_j))>0,
\end{equation}
for every $i\neq j$, where $D_2(\mu\|\nu) \triangleq \sum_x \mu(x)\logtwo(\mu(x)/\nu(x))$ denotes relative entropy on the binary alphabet $\{0,1\}$, the choice of $\delta$ in \eqref{eq:aoi_markov_delta} is feasible. Moreover, the intervals $[p_i-\delta,p_i+\delta]$ are pairwise disjoint, and $\xi$ in \eqref{eq:aoi_markov_xi} is strictly positive because each continuous $g_{ij}$ attains a positive minimum on the compact interval $[p_i-\delta,p_i+\delta]$. Fix any controller $\kappa\in\Pi_B$ and choose a deterministic measurable selector
\begin{equation}
    u_{n-1}^\dagger(z,y)\in \arg\max_{u\in[K]}
    \E_e^\kappa\!\left[Q_e^*(b_{n-1},u)\mid Z_{n-1}=z,Y_{n-1}=y\right]
    \quad \forall (z,y)\in \supp(Z_{n-1},Y_{n-1}),
    \nonumber
\end{equation}
which is possible because both $Z_{n-1}$ and $Y_{n-1}$ take values in finite sets. Define
\begin{equation}
    A_{n-1}^\dagger \triangleq u_{n-1}^\dagger(Z_{n-1},Y_{n-1}).
    \nonumber
\end{equation}
The gap definition at time $n-1$ gives
\begin{equation}
    \mathcal G_{e,n-1}^{\kappa}(B)
    =
    \E_e^\kappa\!\left[V_e^*(b_{n-1})-Q_e^*(b_{n-1},A_{n-1}^\dagger)\right].
    \label{eq:aoi_markov_gap_kappa_local}
\end{equation}
Define
\begin{equation}
    \nonumber
    \vartheta_n \triangleq \frac{1}{1+a_n}.
\end{equation}
Since $a_n<1$, we have $0.5<\vartheta_n$. To construct the decision witness, define the measurable confidence-label map
\begin{equation}
    \nonumber
    \psi_{\mathrm{aoi}}:\Delta(\cS_{\mathrm{aoi}})\to \{0\}\cup[K]
\end{equation}
by
\begin{equation}
    \psi_{\mathrm{aoi}}(b) \triangleq 
    \begin{cases}
        i, & b((i,n))\ge \vartheta_n \text{ for some }i\in[K],\\
        0, & \text{otherwise,}
    \end{cases}
    \nonumber
\end{equation}
which is well defined because $0.5<\vartheta_n$. Write
\begin{equation}
    \Theta_{n-1} \triangleq \psi_{\mathrm{aoi}}(b_{n-1}),
    \qquad
    \cY_{n-1} \triangleq \{(n,0),(n,1)\},
    \qquad
    E_{n-1} \triangleq \{\Theta_{n-1}\in[K],\,Y_{n-1}\in\cY_{n-1}\},
    \nonumber
\end{equation}
and define the action map
\begin{equation}
    a^*_{\mathrm{aoi}}(i,y) \triangleq i,
    \qquad i\in[K],\ y\in\cY_{n-1}.
    \nonumber
\end{equation}
For later use, note that if $b$ is supported on $\{(i,n):i\in[K]\}$, then for every action $u\in[K]$,
\begin{equation}
    Q_e^*(b,u)
    =
    v(n)\,b((u,n))
    +
    \gamma\,\E_e\!\left[V_e^*(b_{n})\mid b_{n-1}=b\right],
    \label{eq:aoi_markov_qstar}
\end{equation}
because the immediate reward at age $n$ is $v(n)\ind\{u=J_{n-1}\}$ and, since the transition and observation kernels are action-independent, the continuation term does not depend on $u$ given $b$. On $E_{n-1}$, the belief $b_{n-1}$ is supported on $\{(i,n):i\in[K]\}$, so \eqref{eq:aoi_markov_qstar} applies. We now verify the decision witness gap condition. If $\Theta_{n-1}=i$, then $b_{n-1}((i,n))\ge \vartheta_n$ by definition of $\psi_{\mathrm{aoi}}$, while every competing label $u\neq i$ satisfies
\begin{equation}
    b_{n-1}((u,n))\le 1-\vartheta_n=\frac{a_n}{1+a_n}.
    \nonumber
\end{equation}
Using \eqref{eq:aoi_markov_qstar}, for every $u\neq i$,
\begin{align}
    V_e^*(b_{n-1})-Q_e^*(b_{n-1},u)
    &=
    Q_e^*(b_{n-1},i)-Q_e^*(b_{n-1},u)
    \notag\\
    &=
    v(n)\bigl(b_{n-1}((i,n))-b_{n-1}((u,n))\bigr)
    \notag\\
    &\ge
    v(n)\left(\vartheta_n-\frac{a_n}{1+a_n}\right)
    =
    v(n)\frac{1-a_n}{1+a_n}.
    \label{eq:aoi_markov_gap_pointwise}
\end{align}
This verifies the witness gap condition with $\eta= v(n)\frac{1-a_n}{1+a_n}$. We next identify a high-confidence decision witness event. Set
\begin{equation}
    F_{n-1}  \triangleq  \{\tau_{n-1}=n\},
    \nonumber
\end{equation}
so $F_{n-1}$ is the observable event that no hidden-label jump occurred during the first $n-1$ transitions. The forced refresh acts only after a slot has already reached age $n$, so it does not affect the joint law of $(J_0,\ldots,J_{n-1},\tau_0,\ldots,\tau_{n-1},Y_0,\ldots,Y_{n-1})$ on $F_{n-1}$. Since $\tau_0=1$ and the label persists with probability $1-\alpha$ at each transition,
\begin{equation}
    \Pp(F_{n-1})=(1-\alpha)^{n-1}.
    \nonumber
\end{equation}
On $F_{n-1}$ one has $J_{n-1}=J_0$, so because $J_0\sim \mathrm{Unif}([K])$,
\begin{equation}
    J_{n-1}\mid F_{n-1} \sim \mathrm{Unif}([K]).
    \label{eq:aoi_markov_uniform_given_fn}
\end{equation}
For each $i\in[K]$, define the event
\begin{equation}
    T_i^{(n)}  \triangleq  \{\,|\bar O_n-p_i|\le \delta\,\},
    \qquad
    \bar O_n \triangleq \frac1n\sum_{s=0}^{n-1} O_s.
    \nonumber
\end{equation}
Set
\begin{equation}
    G_{n-1} \triangleq F_{n-1}\cap \bigcup_{i=1}^K \bigl(\{J_{n-1}=i\}\cap T_i^{(n)}\bigr).
    \nonumber
\end{equation}
Fix $i\in[K]$ and a realization $o_{0:n-1}\in T_i^{(n)}$, and write $r \triangleq \bar o_n\in[p_i-\delta,p_i+\delta]$. Conditional on $F_{n-1}\cap\{J_{n-1}=i\}$, the hidden label is constant and equal to $i$ over the first $n$ slots, so $O_0,\ldots,O_{n-1}$ are i.i.d.\ Bernoulli$(p_i)$. Likewise, conditional on $F_{n-1}\cap\{J_{n-1}=j\}$, they are i.i.d.\ Bernoulli$(p_j)$.
Therefore, for every $j\neq i$,
\begin{align}
    \frac{\Pp(o_{0:n-1}\mid J_{n-1}=i,F_{n-1})}
         {\Pp(o_{0:n-1}\mid J_{n-1}=j,F_{n-1})}
    &=
    \left(\frac{p_i}{p_j}\right)^{\sum_{s=0}^{n-1} o_s}
    \left(\frac{1-p_i}{1-p_j}\right)^{n-\sum_{s=0}^{n-1} o_s}
     \notag\\
    &=
    2^{n(f_{p_i}(r)-f_{p_j}(r))}
    \ge 2^{n\xi}.
    \nonumber
\end{align}
Using \eqref{eq:aoi_markov_uniform_given_fn} and Bayes' rule,
\begin{equation}
    \Pp(J_{n-1}=i\mid O_{0:n-1}=o_{0:n-1},F_{n-1})
    \ge
    \frac{1}{1+(K-1)2^{-n\xi}}
    =
    \vartheta_n.
    \nonumber
\end{equation}
On $F_{n-1}$, the age path is deterministically $(1,2,\ldots,n)$ along times $0,1,\ldots,n-1$, so conditioning on the full observation history $Y_{0:n-1}$ is equivalent to conditioning on $(O_{0:n-1},F_{n-1})$. Hence, on $F_{n-1}\cap\{J_{n-1}=i\}\cap T_i^{(n)}$,
\begin{equation}
    b_{n-1}((i,n))
    =
    \Pp(J_{n-1}=i,\tau_{n-1}=n\mid Y_{0:n-1})
    =
    \Pp(J_{n-1}=i\mid O_{0:n-1},F_{n-1})
    \ge \vartheta_n.
    \nonumber
\end{equation}
Therefore $\Theta_{n-1}=i$ on that event. Since also $Y_{n-1}=(n,O_{n-1})\in\cY_{n-1}$ on $F_{n-1}$, we obtain
\begin{equation}
    G_{n-1}\subseteq E_{n-1},
    \qquad
    \Theta_{n-1}=J_{n-1}\text{ on }G_{n-1}.
    \label{eq:aoi_markov_subset}
\end{equation}
Next, on $E_{n-1}$ the belief $b_{n-1}$ is supported on $\{(i,n):i\in[K]\}$. Since $a_n<1$, the Bayes-optimal action on $E_{n-1}$ is unique and equals $\Theta_{n-1}$. Combining \eqref{eq:aoi_markov_gap_pointwise} with \eqref{eq:aoi_markov_subset} yields the pointwise bound
\begin{equation}
    V_e^*(b_{n-1})-Q_e^*(b_{n-1},A_{n-1}^\dagger)
    \ge
    v(n)\frac{1-a_n}{1+a_n}\,\ind\{A_{n-1}^\dagger\neq J_{n-1}\}\,\ind(G_{n-1}).
    \nonumber
\end{equation}
Taking expectations and using \eqref{eq:aoi_markov_gap_kappa_local} gives
\begin{equation}
    \mathcal G_{e,n-1}^{\kappa}(B)
    \ge
    v(n)\frac{1-a_n}{1+a_n}\,\Pp_e^\kappa(A_{n-1}^\dagger\neq J_{n-1},\ G_{n-1}).
    \label{eq:aoi_markov_reduce_error}
\end{equation}
We next lower-bound the error term. Since $G_{n-1}\subseteq F_{n-1}$,
\begin{align}
    \Pp_e^\kappa(A_{n-1}^\dagger\neq J_{n-1},\ G_{n-1})
    &\ge
    \Pp_e^\kappa(A_{n-1}^\dagger\neq J_{n-1},\ F_{n-1})-\Pp_e^\kappa(G_{n-1}^c,\ F_{n-1}) \notag\\
    &=
    \Pp(F_{n-1})\,\Pp_e^\kappa(A_{n-1}^\dagger\neq J_{n-1}\mid F_{n-1})
    -
    \Pp(F_{n-1})\,\Pp(G_{n-1}^c\mid F_{n-1}).
    \nonumber
\end{align}
By Hoeffding's inequality, under $F_{n-1}\cap\{J_{n-1}=i\}$ the variables $O_0,\ldots,O_{n-1}$ are i.i.d.\ Bernoulli$(p_i)$, so
\begin{equation}
    \Pp(T_i^{(n)c}\mid F_{n-1},J_{n-1}=i)\le 2e^{-2n\delta^2}
    \qquad \forall i\in[K].
    \nonumber
\end{equation}
Averaging over the uniform conditional law of $J_{n-1}$ given $F_{n-1}$ gives
\begin{equation}
    \Pp(G_{n-1}^c\mid F_{n-1})\le 2e^{-2n\delta^2}.
    \nonumber
\end{equation}
Therefore,
\begin{equation}
    \Pp_e^\kappa(A_{n-1}^\dagger\neq J_{n-1},\ G_{n-1})
    \ge
    (1-\alpha)^{n-1}
    \Bigl(p_{\mathrm{err}}-2e^{-2n\delta^2}\Bigr),
    \qquad
    p_{\mathrm{err}} \triangleq \Pp_e^\kappa(A_{n-1}^\dagger\neq J_{n-1}\mid F_{n-1}).
    \label{eq:aoi_markov_error_split_final}
\end{equation}
It remains to lower-bound $p_{\mathrm{err}}$. Write $h_2(u) \triangleq -u\logtwo u-(1-u)\logtwo(1-u)$ for the binary entropy function. Under $F_{n-1}$, the hidden label $J_{n-1}$ is uniform on $[K]$, so
\begin{equation}
    H(J_{n-1}\mid F_{n-1})=\logtwo K.
    \nonumber
\end{equation}
Also, on $F_{n-1}$, the current observation is $Y_{n-1}=(n,O_{n-1})$. Hence, it carries at most one bit:
\begin{equation}
    H(Y_{n-1}\mid F_{n-1})=H(O_{n-1}\mid F_{n-1})\le 1.
    \nonumber
\end{equation}
Therefore,
\begin{align}
    H(J_{n-1}\mid Y_{n-1},F_{n-1})
    &=
    H(J_{n-1}\mid F_{n-1})-I(J_{n-1};Y_{n-1}\mid F_{n-1}) \notag\\
    &\ge
    H(J_{n-1}\mid F_{n-1})-H(Y_{n-1}\mid F_{n-1}) \notag\\
    &\ge
    \logtwo K-1.
    \nonumber
\end{align}
Because $A_{n-1}^\dagger$ is a deterministic function of $(Z_{n-1},Y_{n-1})$,
\begin{equation}
    I(J_{n-1};A_{n-1}^\dagger\mid Y_{n-1},F_{n-1})
    \le I(J_{n-1};Z_{n-1}\mid Y_{n-1},F_{n-1})
    \le B.
    \nonumber
\end{equation}
Thus
\begin{align}
    H(J_{n-1}\mid A_{n-1}^\dagger,Y_{n-1},F_{n-1}) &= H(J_{n-1}\mid Y_{n-1},F_{n-1}) - I(J_{n-1};A_{n-1}^\dagger\mid Y_{n-1},F_{n-1})\notag \\
    &\ge \logtwo K-1-B.
    \label{eq:aoi_markov_postdecode_entropy}    
\end{align}
Applying Fano's inequality conditionally on $F_{n-1}$ gives
\begin{equation}
    H(J_{n-1}\mid A_{n-1}^\dagger,Y_{n-1},F_{n-1})
    \le h_2(p_{\mathrm{err}})+p_{\mathrm{err}}\logtwo(K-1)
    \le 1+p_{\mathrm{err}}\logtwo K.
    \nonumber
\end{equation}
Combining this with \eqref{eq:aoi_markov_postdecode_entropy} yields
\begin{equation}
    p_{\mathrm{err}}
    \ge
    \left(1-\frac{B+2}{\logtwo K}\right)_+.
    \label{eq:aoi_markov_fano}
\end{equation}
Substituting \eqref{eq:aoi_markov_fano} into \eqref{eq:aoi_markov_error_split_final}, using nonnegativity, and then combining with \eqref{eq:aoi_markov_reduce_error}, we obtain
\begin{align}
    \mathcal G_{e,n-1}^{\kappa}(B)
    &\ge
    (1-\alpha)^{n-1}
    \,v(n)\frac{1-a_n}{1+a_n}
    \Bigl(p_{\mathrm{err}}-2e^{-2n\delta^2}\Bigr)_+ \notag\\
    &\ge
    (1-\alpha)^{n-1}
    \,v(n)\frac{1-a_n}{1+a_n}
    \left(1-2e^{-2n\delta^2}-\frac{B+2}{\logtwo K}\right)_+.
    \nonumber
\end{align}
Taking the infimum over $\kappa\in\Pi_B$ and applying the stagewise converse
\begin{equation}
    D_e(B)\ge \gamma^{n-1}\inf_{\kappa\in\Pi_B}\mathcal G_{e,n-1}^{\kappa}(B)    
    \nonumber
\end{equation}
gives \eqref{eq:aoi_markov_frontier}.
\end{proof}

\section{\textsc{RECAP} Performance Bounds}
\label{app:recap_gap}
This appendix provides the definitions and proof for Theorem~\ref{thm:recap_regularized_bound}. In particular, Section~\ref{sec:app_recap_gap_preliminaries} provides the necessary preliminaries used in the proof. Section~\ref{sec:app_recap_gap_bounds} then proves the regularized \textsc{RECAP} performance bound and states the corresponding exact-correction bound under a local tube condition.

\subsection{Preliminaries}
\label{sec:app_recap_gap_preliminaries}
For $\bar b,\bar b'\in \Delta(\cS)$, define
\begin{equation}
    d_{\mathrm{pred}}^e(\bar b,\bar b')
     \triangleq 
    \sup_{y\in \cY}\sup_{a\in\cA}
    \tv{\bar T_a(C_y(\bar b))}{\bar T_a(C_y(\bar b'))}.
    \label{eq:recap_grounded_predictive_metric}
\end{equation}
Thus, $d_{\mathrm{pred}}^e$ measures the largest one-step future predictive discrepancy that the two predictive priors can induce after current-observation correction and one controlled transition. 

\begin{proposition}[Prediction contraction]
\label{prop:recap_grounded_prediction_contraction}
For every action $a\in\cA$ and all beliefs $b,\hat b\in \Delta(\cS)$,
\begin{equation}
    \nonumber
    \tv{\bar T_a(b)}{\bar T_a(\hat b)}\le \tv{b}{\hat b}.
\end{equation}
Consequently,
\begin{equation}
    d_{\mathrm{pred}}^e(\bar b,\bar b')
    \le
    \sup_{y\in \cY}\tv{C_y(\bar b)}{C_y(\bar b')}.
    \label{eq:prediction_contraction_second}
\end{equation}
\end{proposition}

\begin{proof}
The map $b\mapsto \bar T_a(b)$ is the pushforward of $b$ through the one-step state-transition kernel under action $a$. Total variation is contractive under stochastic kernels, so
\begin{equation}
    \nonumber
    \tv{\bar T_a(b)}{\bar T_a(\hat b)}\le \tv{b}{\hat b}.
\end{equation}
Applying this pointwise inside \eqref{eq:recap_grounded_predictive_metric} gives \eqref{eq:prediction_contraction_second}.
\end{proof}

\begin{proposition}[Action-value perturbation bounds under total variation]
\label{prop:recap_grounded_Q_tv}
For every finite POMDP, all beliefs $b,\hat b\in\Delta(\cS)$, and every action $a\in\cA$,
\begin{equation}
    \nonumber
    \bigl|Q_e^*(b,a)-Q_e^*(\hat b,a)\bigr|
    \le
    \frac{2R_{\max}\tv{b}{\hat b}}{1-\gamma}.
\end{equation}
Consequently, for every tie-broken greedy action $a^*(\hat b)$ at $\hat b$,
\begin{equation}
    \label{eq:recap_grounded_Q_tv_second}
    V_e^*(b)-Q_e^*(b,a^*(\hat b))
    \le
    \frac{4R_{\max}\tv{b}{\hat b}}{1-\gamma}.
\end{equation}
\end{proposition}

\begin{proof}
Fix an action $a\in\cA$. Let $\Pi_{\mathrm{hist}}(a)$ denote the class of admissible history-based continuation policies used from time $1$ onward after taking action $a$ at time $0$. For any $\pi\in\Pi_{\mathrm{hist}}(a)$, define
\begin{equation}
    \nonumber
    q_e^{\pi}(s,a)
    =
    \E_e^{\pi}\!\left[\sum_{t=0}^{\infty}\gamma^t r_e(S_t,A_t)\,\middle|\, S_0=s,\ A_0=a\right],
\end{equation}
so that $\lvert q_e^{\pi}(s,a)\rvert\le R_{\max}/(1-\gamma)$ for every state $s$. Since the continuation policy and the first action are fixed, the initial hidden state enters linearly, and therefore
\begin{equation}
    \nonumber
    Q_e^{\pi}(b,a)=\sum_{s\in\cS} b(s)\,q_e^{\pi}(s,a)
    \qquad \forall b\in\Delta(\cS).
\end{equation}
Optimizing over continuation policies from time $1$ onward yields
\begin{equation}
    \nonumber
    Q_e^*(b,a)=\sup_{\pi\in\Pi_{\mathrm{hist}}(a)} Q_e^{\pi}(b,a).
\end{equation}
Hence
\begin{align*}
    Q_e^*(b,a)-Q_e^*(\hat b,a)
    &\le \sup_{\pi\in\Pi_{\mathrm{hist}}(a)}\bigl(Q_e^{\pi}(b,a)-Q_e^{\pi}(\hat b,a)\bigr)
    \\
    &= \sup_{\pi\in\Pi_{\mathrm{hist}}(a)}\sum_{s\in\cS}(b(s)-\hat b(s))q_e^{\pi}(s,a)
    \\
    &\le \sup_{\pi\in\Pi_{\mathrm{hist}}(a)}\|q_e^{\pi}(\cdot,a)\|_{\infty}\,\|b-\hat b\|_1
    \\
    &\le \frac{2R_{\max}}{1-\gamma}\,\tv{b}{\hat b}.
\end{align*}
Interchanging $b$ and $\hat b$ gives the fixed-action perturbation bound. For \eqref{eq:recap_grounded_Q_tv_second}, add and subtract the greedy action at $\hat b$:
\begin{align*}
    V_e^*(b)-Q_e^*(b,a^*(\hat b))
    &=
    Q_e^*(b,a^*(b))-Q_e^*(b,a^*(\hat b))
    \\
    &\le
    \bigl(Q_e^*(b,a^*(b))-Q_e^*(\hat b,a^*(b))\bigr)
    +
    \bigl(Q_e^*(\hat b,a^*(\hat b))-Q_e^*(b,a^*(\hat b))\bigr),
\end{align*}
where the inequality uses the optimality of $a^*(\hat b)$ at $\hat b$. Applying the fixed-action bound to both differences gives the factor $4R_{\max}/(1-\gamma)$.
\end{proof}

\begin{proposition}[Correction stability under a likelihood margin]
\label{prop:recap_grounded_update_lipschitz_sufficient}
Fix an observation $y\in\cY$. Suppose $\bar b,\bar b'\in \Delta(\cS)$ satisfy
\begin{equation}
    \nonumber
    p_e(y\mid \bar b)\ge p_{\min},
    \qquad
    p_e(y\mid \bar b')\ge p_{\min}
\end{equation}
for some $p_{\min}>0$. Then
\begin{equation}
    \nonumber
    \tv{C_y(\bar b)}{C_y(\bar b')}
    \le
    \frac{2}{p_{\min}}\,\tv{\bar b}{\bar b'}.
\end{equation}
\end{proposition}

\begin{proof}
Let
\begin{equation}
\nonumber
    u(s)\triangleq u_y(\bar b)(s)=O_e(y\mid s)\bar b(s),
    \qquad
    v(s)\triangleq u_y(\bar b')(s)=O_e(y\mid s)\bar b'(s).
\end{equation}
Then
\begin{equation}
    \nonumber
    C_y(\bar b)=\frac{u}{\|u\|_1},
    \qquad
    C_y(\bar b')=\frac{v}{\|v\|_1},
\end{equation}
and
\begin{equation}
    \nonumber
    \|u\|_1=p_e(y\mid \bar b)\ge p_{\min},
    \qquad
    \|v\|_1=p_e(y\mid \bar b')\ge p_{\min}.
\end{equation}
First, since $0\le O_e(y\mid s)\le 1$ for every $s\in\cS$,
\begin{equation}
    \|u-v\|_1
    =
    \sum_{s\in\cS} O_e(y\mid s)\,|\bar b(s)-\bar b'(s)|
    \le
    \sum_{s\in\cS} |\bar b(s)-\bar b'(s)|
    =
    \|\bar b-\bar b'\|_1.
\label{eq:b_bound}
\end{equation}
Next, write
\begin{equation}
    \frac{u}{\|u\|_1}-\frac{v}{\|v\|_1}
    =
    \frac{u-v}{\|u\|_1}
    +
    v\left(\frac{1}{\|u\|_1}-\frac{1}{\|v\|_1}\right).
    \nonumber
\end{equation}
Taking $\ell_1$ norms and using the triangle inequality gives
\begin{equation}
    \nonumber
    \left\|\frac{u}{\|u\|_1}-\frac{v}{\|v\|_1}\right\|_1
    \le
    \frac{\|u-v\|_1}{\|u\|_1}
    +
    \|v\|_1\left|\frac{1}{\|u\|_1}-\frac{1}{\|v\|_1}\right|.
\end{equation}
For the second term,
\begin{equation}
    \|v\|_1\left|\frac{1}{\|u\|_1}-\frac{1}{\|v\|_1}\right|
    =
    \frac{|\|u\|_1-\|v\|_1|}{\|u\|_1}
    \le
    \frac{\|u-v\|_1}{\|u\|_1},
    \nonumber
\end{equation}
where we used
\begin{equation}
    \nonumber
    |\|u\|_1-\|v\|_1|\le \|u-v\|_1.
\end{equation}
Therefore
\begin{equation}
    \nonumber
    \left\|\frac{u}{\|u\|_1}-\frac{v}{\|v\|_1}\right\|_1
    \le
    \frac{2}{\|u\|_1}\|u-v\|_1
    \le
    \frac{2}{p_{\min}}\|u-v\|_1.
\end{equation}
Combining with \eqref{eq:b_bound} yields
\begin{equation}
    \nonumber
    \left\|\frac{u}{\|u\|_1}-\frac{v}{\|v\|_1}\right\|_1
    \le
    \frac{2}{p_{\min}}\|\bar b-\bar b'\|_1.
\end{equation}
Finally, dividing by $2$ and using $\tv{\mu}{\nu}=\frac12\|\mu-\nu\|_1$ gives
\begin{equation}
    \nonumber
    \tv{C_y(\bar b)}{C_y(\bar b')}
    \le
    \frac{2}{p_{\min}}\,\tv{\bar b}{\bar b'}.
\end{equation}
\end{proof}

For the regularized correction, fix the full-support anchor $\nu\in\Delta(\cS)$ from Definition~\ref{def:recap_grounded_structural}. We use the shorthand
\begin{equation}
    S_\delta(\bar b) \triangleq (1-\delta)\bar b+\delta\nu,
    \qquad
    m_\nu(y) \triangleq \sum_{s\in\cS}\nu(s)O_e(y\mid s),
    \nonumber
\end{equation}
and
\begin{equation}
    u_y(\bar b)(s) \triangleq O_e(y\mid s)\bar b(s).
    \nonumber
\end{equation}
Then the regularized correction in Definition~\ref{def:recap_grounded_structural}
can be written as $C_{y,\delta}(\bar b)=C_y(S_\delta(\bar b))$ for
$\delta>0$.

\begin{proposition}[Properties of the regularized members $\{C_{y,\delta}\}_{\delta>0}$]
\label{prop:recap_grounded_anchor_regularized_family}
Fix $\nu$ and $\delta\in(0,1]$ as in Definition~\ref{def:recap_grounded_structural}. Then:
\begin{enumerate}[label=(\roman*)]
\item for every $y\in\cY$ and all $\bar b,\bar b'\in\Delta(\cS)$,
\begin{equation}
    \tv{C_{y,\delta}(\bar b)}{C_{y,\delta}(\bar b')}
    \le \Lambda_{y,\delta}\,\tv{\bar b}{\bar b'},
    \nonumber
\end{equation}
where
\begin{equation}
    \nonumber
    \Lambda_{y,\delta}
     \triangleq 
    \begin{cases}
        \dfrac{2(1-\delta)}{\delta\,m_\nu(y)}, & \text{if }m_\nu(y)>0,\\
        0, & \text{if }m_\nu(y)=0,
    \end{cases}
    \qquad
    \Lambda_{\delta} \triangleq \max_{y\in\cY}\Lambda_{y,\delta}<\infty;
\end{equation}
\item defining, for each $y\in\cY$,
\begin{equation}
    \nonumber
    \beta_{y,\delta}
     \triangleq 
    \begin{cases}
        \sup\limits_{\substack{
        \bar b\in\Delta(\cS):\\
        p_e(y\mid \bar b)>0
        }}
        \tv{C_{y,0}(\bar b)}{C_{y,\delta}(\bar b)}, & \exists \bar b\in\Delta(\cS)\text{ with }p_e(y\mid \bar b)>0,\\
        0, & \text{otherwise},
    \end{cases}
\end{equation}
and
\begin{equation}
    \nonumber
    \beta_{\delta} \triangleq \max_{y\in\cY}\beta_{y,\delta},
\end{equation}
one has $0\le \beta_{\delta}\le 1$; moreover, for every fixed $y\in\cY$ and every fixed $\bar b\in\Delta(\cS)$ with $p_e(y\mid \bar b)>0$,
\begin{equation}
    \tv{C_{y,0}(\bar b)}{C_{y,\delta}(\bar b)}\to 0
    \qquad\text{as }\delta\to 0;
    \nonumber
\end{equation}
\item for every $\delta\in(0,1]$, $\Lambda_\delta<\infty$, and in particular $\Lambda_1=0$.
\end{enumerate}
\end{proposition}

\begin{proof}
For item (i), fix $y\in\cY$. If $m_\nu(y)=0$, then $C_{y,\delta}(\cdot)\equiv \nu$ is constant and the claim is trivial. If $m_\nu(y)>0$, then for every $\bar b\in\Delta(\cS)$,
\begin{equation}
    \nonumber
    p_e\bigl(y\mid S_\delta(\bar b)\bigr)=(1-\delta)p_e(y\mid \bar b)+\delta m_\nu(y)\ge \delta m_\nu(y)>0.
\end{equation}
Applying Proposition~\ref{prop:recap_grounded_update_lipschitz_sufficient} to the two priors $S_\delta(\bar b)$ and $S_\delta(\bar b')$ therefore gives
\begin{equation}
    \nonumber
    \tv{C_{y,\delta}(\bar b)}{C_{y,\delta}(\bar b')}
    \le \frac{2}{\delta m_\nu(y)}\tv{S_\delta(\bar b)}{S_\delta(\bar b')}
    =\frac{2(1-\delta)}{\delta m_\nu(y)}\tv{\bar b}{\bar b'}.
\end{equation}
This proves item (i). For item (ii), each quantity inside the defining supremum for $\beta_{y,\delta}$ is a total-variation distance between two probability measures on $\cS$, and therefore lies in $[0,1]$. The fallback value on globally impossible observations is also $0$. Hence, $0\le \beta_{\delta}\le 1$. Fix $y\in\cY$ and $\bar b\in\Delta(\cS)$ with $p_e(y\mid \bar b)>0$. Then $m_\nu(y)>0$ as well because $\nu$ has full support, so Definition~\ref{def:recap_grounded_structural} gives $C_{y,\delta}(\bar b)=C_y(S_\delta(\bar b))$. Now $S_\delta(\bar b)\to \bar b$ coordinatewise, the numerator $u_y(S_\delta(\bar b))(s)=O_e(y\mid s)S_\delta(\bar b)(s)$ converges coordinatewise to $u_y(\bar b)(s)$, and the denominator $p_e(y\mid S_\delta(\bar b))=(1-\delta)p_e(y\mid \bar b)+\delta m_\nu(y)$ converges to the positive value $p_e(y\mid \bar b)$. Hence, $C_{y,\delta}(\bar b)\to C_y(\bar b)=C_{y,0}(\bar b)$ in total variation. Item (iii) follows directly from the definition of $\Lambda_\delta$, since $\cY$ is finite and $\Lambda_1=0$.
\end{proof}

\paragraph*{Why this regularizer is used.}
For $\delta>0$, the map $C_{y,\delta}$ applies exact Bayes after prior smoothing
$S_\delta(\bar b)=(1-\delta)\bar b+\delta\nu$. It preserves the observation
dependence of the update and converges pointwise to exact Bayes on every fixed
positive-likelihood observation as $\delta\to 0$. The constants
$\Lambda_\delta$ and $\beta_\delta$ quantify, respectively, the correction
sensitivity and the smoothing bias. The closed-form bound below requires the induced scalar error recursion to be discount-stable. The exact-correction bound for $\delta=0$ instead uses the constant $\Lambda$ of the following local tube condition.

\begin{definition}[Tube Condition for $\delta=0$]
\label{def:recap_grounded_tube_condition}
Fix a budget-feasible \textsc{RECAP} architecture $(\cZ,\mathcal{C},\operatorname{enc},\pi_{0},\phi_0)$ obtained
from Definition~\ref{def:recap_grounded_structural} at $\delta=0$ by using the
decision rule $\widehat a_H$. The subscript $0$ on $\pi_0$ and $\phi_0$ denotes the controller induced by the exact correction map $C_{y,0}$, not the time index. Write
\begin{equation}
    \nonumber
    z_0 \triangleq \operatorname{enc}(\xi_e),
    \qquad
    \bar e_{0,0} \triangleq \tv{\xi_e}{\bar c_{z_0}},
\end{equation}
and
\begin{equation}
    \nonumber
    \eta_0(z,y)
     \triangleq 
    \tv{\bar T_{\pi_0(z,y)}(C_{y,0}(\bar c_{z}))}
    {\bar c_{\phi_0(z,y,\pi_0(z,y))}}.
\end{equation}
We call a pair $(\rho,\Lambda)$, where $\rho:\cZ\to [0,1]$ and
$\Lambda\in[0,\infty)$, a \emph{feasible tube} if
\begin{enumerate}[label=(\roman*)]
\item $\rho(z_0)\ge \bar e_{0,0}$;
\item for every $z\in \cZ$ and every $y\in\cY$, let
\begin{equation}
    \nonumber
    z'_{z,y} \triangleq \phi_0(z,y,\pi_0(z,y)).
\end{equation}
If
\begin{equation}
    \nonumber
    p_e(y\mid \bar c_z)+\rho(z)=0,
\end{equation}
then no further constraint is imposed, whereas otherwise one
requires
\begin{equation}
    \nonumber
    p_e(y\mid \bar c_z)-\rho(z)>0,
    \qquad
    \Lambda\ge \frac{2}{p_e(y\mid \bar c_z)-\rho(z)},
\end{equation}
and
\begin{equation}
    \nonumber
    \rho\bigl(z'_{z,y}\bigr)\ge \Lambda\rho(z)+\eta_0(z,y).
\end{equation}
\end{enumerate}
\end{definition}

\begin{proposition}[Tube stability bound]
\label{prop:recap_grounded_explicit_tube_condition}
Fix a budget-feasible exact \textsc{RECAP} architecture $(\cZ,\mathcal{C},\operatorname{enc},\pi_0,\phi_0)$ and
let $(\rho,\Lambda)$ be a feasible tube in the sense of
Definition~\ref{def:recap_grounded_tube_condition}. Then along the closed-loop
process induced by the corresponding candidate controller,
\begin{equation}
    \tv{C_{Y_t,0}(\bar b_t)}{C_{Y_t,0}(\bar c_{Z_t})}
    \le
    \Lambda\,\tv{\bar b_t}{\bar c_{Z_t}}
    \qquad\text{a.s. } \forall t\ge 0.
    \nonumber
\end{equation}
Moreover,
\begin{equation}
    \nonumber
    \tv{\bar b_t}{\bar c_{Z_t}}\le \rho(Z_t) \qquad \text{a.s. } \forall t\ge 0,
\end{equation}
and therefore
\begin{equation}
    \nonumber
    \tv{C_{Y_t,0}(\bar b_t)}{C_{Y_t,0}(\bar c_{Z_t})}
    \le \Lambda\rho(Z_t).
\end{equation}
\end{proposition}

\begin{proof}
Let $(Z_t)_{t\ge0}$ denote the closed-loop memory process generated by the
candidate:
\begin{equation}
    \nonumber
    Z_0 \triangleq \operatorname{enc}(\xi_e),
    \qquad
    A_t \triangleq \pi_0(Z_t,Y_t),
    \qquad
    Z_{t+1} \triangleq \phi_0(Z_t,Y_t,A_t).
\end{equation}
By item~(i) of Definition~\ref{def:recap_grounded_tube_condition},
\begin{equation}
    \nonumber
    \tv{\bar b_0}{\bar c_{Z_0}}
    =
    \tv{\xi_e}{\bar c_{\operatorname{enc}(\xi_e)}}
    \le \rho(Z_0).
\end{equation}
Assume inductively that $\tv{\bar b_t}{\bar c_{Z_t}}\le \rho(Z_t)$. Fix $Z_t=z$ and $Y_t=y$. Since total variation controls event probabilities,
\begin{equation}
    \nonumber
    \bigl|p_e(y\mid \bar b_t)-p_e(y\mid \bar c_z)\bigr|
    \le \tv{\bar b_t}{\bar c_z}
    \le \rho(z).
\end{equation}
If $p_e(y\mid \bar c_z)+\rho(z)=0$, then $p_e(y\mid \bar b_t)=0$, so that
observation cannot occur under the induction hypothesis. Therefore every realized
observation satisfies the second case in
Definition~\ref{def:recap_grounded_tube_condition}(ii), and hence
\begin{equation}
    \nonumber
    p_e(y\mid \bar b_t)\ge p_e(y\mid \bar c_z)-\rho(z)>0.
\end{equation}
Proposition~\ref{prop:recap_grounded_update_lipschitz_sufficient} then gives
\begin{equation}
    \nonumber
    \tv{C_{y,0}(\bar b_t)}{C_{y,0}(\bar c_z)}
    \le
    \frac{2}{p_e(y\mid \bar c_z)-\rho(z)}
    \tv{\bar b_t}{\bar c_z}
    \le \Lambda\,\tv{\bar b_t}{\bar c_z}.
\end{equation}
Applying Proposition~\ref{prop:recap_grounded_prediction_contraction} and then
Definition~\ref{def:recap_grounded_tube_condition}(ii) again,
\begin{align}
    \tv{\bar b_{t+1}}{\bar c_{Z_{t+1}}}
    &=
    \tv{\bar T_{\pi_0(z,y)}(C_{y,0}(\bar b_t))}
    {\bar c_{\phi_0(z,y,\pi_0(z,y))}} \nonumber \\
    &\le
    \tv{C_{y,0}(\bar b_t)}{C_{y,0}(\bar c_z)}
    +\eta_0(z,y)
    \nonumber \\
    &\le
    \Lambda\rho(z)+\eta_0(z,y)
    \nonumber \\
    &\le
    \rho\bigl(\phi_0(z,y,\pi_0(z,y))\bigr).\nonumber
\end{align}
This closes the induction and proves the claim.
\end{proof}

\paragraph*{The truncated planner.}
For $h=0,1,\dots,H$, define recursively
\begin{equation}
    \nonumber
    \widehat V_0(b) \triangleq 0,
\end{equation}
and for $h\ge 1$,
\begin{equation}
    \nonumber
    \widehat Q_h(b,a)
     \triangleq 
    \sum_{s\in \cS} b(s)\,r_e(s,a)
    +
    \gamma \sum_{y\in \cY} p_e(y\mid \bar T_a(b))\,\widehat V_{h-1}(C_y(\bar T_a(b))),
\end{equation}
\begin{equation}
    \nonumber
    \widehat V_h(b) \triangleq \max_{a\in \cA}\widehat Q_h(b,a).
\end{equation}
We use the exact $H$-step truncated action-value function $\widehat Q_H$. Define the planner-greedy action and planner-induced gap on all beliefs by
\begin{equation}
    \nonumber
    \widehat a_H(b)\in \operatorname*{argmax}_{a\in \cA}\widehat Q_H(b,a),
\end{equation}
where the global deterministic tie-breaking convention resolves ties, and define
\begin{equation}
    \nonumber
    \widehat g_H(b)
     \triangleq 
    \max_{a\in \cA}\widehat Q_H(b,a)
    -
    \max_{a\neq \widehat a_H(b)}\widehat Q_H(b,a).
\end{equation}
Because $0\le \widehat Q_H(b,a)\le R_{\max}/(1-\gamma)$ for every $(b,a)$, one has
\begin{equation}
    \nonumber
    0\le \widehat g_H(b)\le \frac{R_{\max}}{1-\gamma}\le \frac{2R_{\max}}{1-\gamma}.
\end{equation}
In Theorem~\ref{thm:recap_regularized_bound} and the exact-correction bound, the $\delta$-indexed \textsc{RECAP} construction of Definition~\ref{def:recap_grounded_structural} uses the strict truncated-planner rule $\widehat a_H$.

\begin{proposition}[Uniform truncation error of the strict planner]
\label{prop:recap_grounded_qhat_gap_correctness}
For every $b\in \Delta(\cS)$ and $a\in \cA$,
\begin{equation}
    \nonumber
    0\le Q_e^*(b,a)-\widehat Q_H(b,a)\le \varepsilon_H,
    \qquad
    \varepsilon_H \triangleq \frac{\gamma^H R_{\max}}{1-\gamma}.
\end{equation}
Consequently,
\begin{equation}
    \nonumber
    \sup_{b\in \Delta(\cS)}\sup_{a\in \cA}
    \bigl|Q_e^*(b,a)-\widehat Q_H(b,a)\bigr|
    \le \varepsilon_H.
\end{equation}
\end{proposition}

\begin{proof}
A standard finite-horizon dynamic-programming induction shows that $\widehat Q_H$ is the exact optimal $H$-stage truncated action-value function. Because rewards are nonnegative, truncation can only decrease value, so $\widehat Q_H(b,a)\le Q_e^*(b,a)$. The omitted tail after stage $H-1$ is uniformly bounded by
\begin{equation}
    \nonumber
    \sum_{t=H}^{\infty}\gamma^t R_{\max}
    =
    \frac{\gamma^H R_{\max}}{1-\gamma}.
\end{equation}
\end{proof}

\subsection{Performance Bounds}
\label{sec:app_recap_gap_bounds}

\paragraph*{Regularized-correction bound.}
For $\delta\in(0,1]$, suppose $\gamma\Lambda_\delta<1$. Define
\begin{equation}
    \nonumber
    \bar e_{0,\mathcal C}
    \triangleq
    \tv{\xi_e}{\bar c_{\operatorname{enc}_{\mathcal C}(\xi_e)}}
\end{equation}
and
\begin{equation}
    \nonumber
    \bar\eta_{\mathcal C,\delta}
    \triangleq
    \sup_{z\in\{1,\ldots,K\}}
    \sup_{y\in\cY}
    \tv{
    \bar T_{\widehat a_H(C_{y,\delta}(\bar c_z))}
    \!\left(C_{y,\delta}(\bar c_z)\right)
    }{
    \bar c_{
    \operatorname{enc}_{\mathcal C}\left(
    \bar T_{\widehat a_H(C_{y,\delta}(\bar c_z))}
    \!\left(C_{y,\delta}(\bar c_z)\right)
    \right)}
    }.
\end{equation}
Then the \textsc{RECAP} controller $\kappa_{\mathcal C,\delta}$ satisfies
\begin{equation}
V_e^*-V_e^{\kappa_{\mathcal C,\delta}}
    \le
    \frac{2\varepsilon_H}{1-\gamma}
    +
    \frac{4R_{\max}}{1-\gamma}
    \left[
    \frac{\beta_\delta}{1-\gamma}
    +
    \Lambda_\delta
    \left(
    \frac{\bar e_{0,\mathcal C}}{1-\gamma\Lambda_\delta}
    +
    \frac{\gamma(\beta_\delta+\bar\eta_{\mathcal C,\delta})}
    {(1-\gamma)(1-\gamma\Lambda_\delta)}
    \right)
    \right].
    \nonumber
\end{equation}

\begin{proof}[Proof of Theorem~\ref{thm:recap_regularized_bound}.]
Let $(Z_t)_{t\ge 0}$ denote the actual FSC memory process generated by
$\kappa_{\mathcal C,\delta}$, i.e.
\begin{equation}
    \nonumber
    Z_0 \triangleq \operatorname{enc}_{\mathcal C}(\xi_e),
    \qquad
    A_t \triangleq \widehat a_H(C_{Y_t,\delta}(\bar c_{Z_t})),
    \qquad
    Z_{t+1} \triangleq 
    \operatorname{enc}_{\mathcal C}\!\left(
    \bar T_{A_t}(C_{Y_t,\delta}(\bar c_{Z_t}))
    \right).
\end{equation}
Write
\begin{equation}
    \nonumber
    \widehat b_{t,\delta} \triangleq C_{Y_t,\delta}(\bar c_{Z_t}).
\end{equation}
Define
\begin{equation}
    \nonumber
    \bar e_{t,\delta} \triangleq \tv{\bar b_t}{\bar c_{Z_t}},
    \qquad
    d_{t,\delta} \triangleq \tv{b_t}{\widehat b_{t,\delta}}.
\end{equation}
By the discounted performance-difference identity together with belief-MDP
equivalence,
\begin{equation}
    \nonumber
    V_e^*-V_e^{\kappa_{\mathcal C,\delta}}
    =
    \E_e^{\kappa_{\mathcal C,\delta}}
    \left[
    \sum_{t=0}^{\infty}\gamma^t
    \bigl(V_e^*(b_t)-Q_e^*(b_t,A_t)\bigr)
    \right].
\end{equation}
Because $A_t=\widehat a_H(\widehat b_{t,\delta})$, Proposition~\ref{prop:recap_grounded_qhat_gap_correctness} gives
\begin{equation}
    \nonumber
    V_e^*(\widehat b_{t,\delta})
    -
    Q_e^*(\widehat b_{t,\delta},A_t)
    \le 2\varepsilon_H.
\end{equation}
Also, since $V_e^*(b)=\max_a Q_e^*(b,a)$, one has
\begin{equation}
    \nonumber
    |V_e^*(b_t)-V_e^*(\widehat b_{t,\delta})|
    \le
    \sup_{a\in\cA}|Q_e^*(b_t,a)-Q_e^*(\widehat b_{t,\delta},a)|.
\end{equation}
Applying Proposition~\ref{prop:recap_grounded_Q_tv} yields
\begin{equation}
    \nonumber
    |V_e^*(b_t)-V_e^*(\widehat b_{t,\delta})|
    \le
    \frac{2R_{\max}}{1-\gamma}d_{t,\delta},
    \qquad
    |Q_e^*(b_t,A_t)-Q_e^*(\widehat b_{t,\delta},A_t)|
    \le
    \frac{2R_{\max}}{1-\gamma}d_{t,\delta}.
\end{equation}
Hence
\begin{equation}
    \nonumber
    V_e^*(b_t)-Q_e^*(b_t,A_t)
    \le
    2\varepsilon_H+\frac{4R_{\max}}{1-\gamma}d_{t,\delta}.
\end{equation}
On every realization, $p_e(Y_t\mid \bar b_t)>0$ and the exact Bayes
posterior $b_t=C_{Y_t,0}(\bar b_t)$ is well defined. Proposition~\ref{prop:recap_grounded_anchor_regularized_family} yields
\begin{align}
    d_{t,\delta}
    &=
    \tv{C_{Y_t,0}(\bar b_t)}{C_{Y_t,\delta}(\bar c_{Z_t})} \notag\\
    &\le
    \tv{C_{Y_t,0}(\bar b_t)}{C_{Y_t,\delta}(\bar b_t)}
    +
    \tv{C_{Y_t,\delta}(\bar b_t)}{C_{Y_t,\delta}(\bar c_{Z_t})}
    \notag\\
    &\le
    \beta_{\delta}+\Lambda_{\delta}\bar e_{t,\delta}. 
    \nonumber
\end{align}
Using Proposition~\ref{prop:recap_grounded_prediction_contraction} together
with the actual FSC update, we obtain
\begin{align}
    \bar e_{t+1,\delta}
    &=
    \tv{\bar T_{A_t}(b_t)}{\bar c_{Z_{t+1}}}\nonumber\\
    &\le
    \tv{\bar T_{A_t}(b_t)}{\bar T_{A_t}(\widehat b_{t,\delta})}
    +
    \tv{\bar T_{A_t}(\widehat b_{t,\delta})}{\bar c_{Z_{t+1}}}
    \nonumber\\
    &\le
    d_{t,\delta}+\bar\eta_{\mathcal C,\delta}
    \nonumber\\
    &\le
    \beta_{\delta}+\Lambda_{\delta}\bar e_{t,\delta}+\bar\eta_{\mathcal C,\delta}.
    \nonumber
\end{align}
Since $\bar e_{0,\delta}=\bar e_{0,\mathcal C}$ and $\gamma\Lambda_\delta<1$, multiplying the recursion by $\gamma^t$ and summing the resulting geometric series gives
\begin{equation}
    \nonumber
    \sum_{t=0}^{\infty}\gamma^t \bar e_{t,\delta}
    \le
    \frac{\bar e_{0,\mathcal C}}{1-\gamma\Lambda_{\delta}}
    +
    \frac{\gamma\bigl(\beta_{\delta}+\bar\eta_{\mathcal C,\delta}\bigr)}
    {(1-\gamma)(1-\gamma\Lambda_{\delta})}.
\end{equation}
Combining this with
\begin{equation}
    \nonumber
    \sum_{t=0}^{\infty}\gamma^t d_{t,\delta}
    \le
    \frac{\beta_\delta}{1-\gamma}
    +
    \Lambda_\delta
    \sum_{t=0}^{\infty}\gamma^t\bar e_{t,\delta}
\end{equation}
proves the displayed bound. The bound on $D_e(B)$ follows because
$\kappa_{\mathcal C,\delta}$ has $K\le 2^B$ memory states and therefore belongs
to $\Pi_B$.
\end{proof}

\paragraph*{Exact-correction bound.}
For $\delta=0$, suppose the local tube condition holds with feasible tube
$(\rho,\Lambda)$ and $\gamma\Lambda<1$. Define
\begin{equation}
    \nonumber
    \bar e_{0,\mathcal C} \triangleq \tv{\xi_e}{\bar c_{\operatorname{enc}_{\mathcal C}(\xi_e)}}
\end{equation}
and
\begin{equation}
    \nonumber
    \bar\eta_{\mathcal C,0}
     \triangleq 
    \sup_{z\in\{1,\ldots,K\}}
    \sup_{y\in\cY}
    \tv{
    \bar T_{\widehat a_H(C_{y,0}(\bar c_z))}
    \!\left(C_{y,0}(\bar c_z)\right)
    }{
    \bar c_{
    \operatorname{enc}_{\mathcal C}\left(
    \bar T_{\widehat a_H(C_{y,0}(\bar c_z))}
    \!\left(C_{y,0}(\bar c_z)\right)
    \right)}
    }.
\end{equation}
Then
\begin{equation}
    V_e^*-V_e^{\kappa_{\mathcal C,0}}
    \le
    \frac{2\varepsilon_H}{1-\gamma}
    +
    \frac{4R_{\max}\Lambda}{1-\gamma}
    \left(
    \frac{\bar e_{0,\mathcal C}}{1-\gamma\Lambda}
    +
    \frac{\gamma\bar\eta_{\mathcal C,0}}
    {(1-\gamma)(1-\gamma\Lambda)}
    \right).
    \nonumber
\end{equation}

\begin{proof}[Proof for $\delta=0$.]
Let $(Z_t)_{t\ge 0}$ denote the actual FSC memory process generated by
$\kappa_{\mathcal C,0}$, i.e.
\begin{equation}
    \nonumber
    Z_0 \triangleq \operatorname{enc}_{\mathcal C}(\xi_e),
    \qquad
    A_t \triangleq \widehat a_H(C_{Y_t,0}(\bar c_{Z_t})),
    \qquad
    Z_{t+1} \triangleq 
    \operatorname{enc}_{\mathcal C}\!\left(
    \bar T_{A_t}(C_{Y_t,0}(\bar c_{Z_t}))
    \right).
\end{equation}
Define also
\begin{equation}
    \nonumber
    \widehat b_{t,0} \triangleq C_{Y_t,0}(\bar c_{Z_t}),
\end{equation}
and
\begin{equation}
    \nonumber
    \bar e_{t,0} \triangleq \tv{\bar b_t}{\bar c_{Z_t}},
    \qquad
    d_{t,0} \triangleq \tv{b_t}{\widehat b_{t,0}}.
\end{equation}
By the discounted performance-difference identity together with belief-MDP
equivalence,
\begin{equation}
V_e^*-V_e^{\kappa_{\mathcal C,0}}
=
\E_e^{\kappa_{\mathcal C,0}}
\left[
\sum_{t=0}^{\infty}\gamma^t
\bigl(V_e^*(b_t)-Q_e^*(b_t,A_t)\bigr)
\right].
\nonumber
\end{equation}
Because $A_t=\widehat a_H(\widehat b_{t,0})$, Proposition~\ref{prop:recap_grounded_qhat_gap_correctness} implies
\begin{equation}
V_e^*(\widehat b_{t,0})-Q_e^*(\widehat b_{t,0},A_t)
\le 2\varepsilon_H.
\nonumber
\end{equation}
Proposition~\ref{prop:recap_grounded_Q_tv}, together with
$V_e^*(b)=\max_{a\in\cA}Q_e^*(b,a)$, gives
\begin{equation}
\nonumber
|V_e^*(b_t)-V_e^*(\widehat b_{t,0})|
\le \frac{2R_{\max}}{1-\gamma}d_{t,0},
\qquad
|Q_e^*(b_t,A_t)-Q_e^*(\widehat b_{t,0},A_t)|
\le \frac{2R_{\max}}{1-\gamma}d_{t,0}.
\end{equation}
Therefore
\begin{equation}
\nonumber
V_e^*(b_t)-Q_e^*(b_t,A_t)
\le
2\varepsilon_H+\frac{4R_{\max}}{1-\gamma}d_{t,0}.
\end{equation}
Taking discounted expectations yields
\begin{equation}
\nonumber
V_e^*-V_e^{\kappa_{\mathcal C,0}}
\le
\frac{2\varepsilon_H}{1-\gamma}
+
\frac{4R_{\max}}{1-\gamma}
\sum_{t=0}^{\infty}\gamma^t\E[d_{t,0}].
\end{equation}
By Proposition~\ref{prop:recap_grounded_explicit_tube_condition},
\begin{equation}
\nonumber
d_{t,0}\le \Lambda\bar e_{t,0}
\qquad\text{almost surely for every }t\ge0.
\end{equation}
Moreover,
\begin{equation}
\bar b_{t+1}=\bar T_{A_t}(b_t),
\qquad
Z_{t+1}=\operatorname{enc}_{\mathcal C}\!\left(
\bar T_{A_t}(C_{Y_t,0}(\bar c_{Z_t}))
\right),
\nonumber
\end{equation}
so prediction contraction and the one-step reprojection definition give
\begin{align*}
\bar e_{t+1,0}
&\le
\tv{\bar T_{A_t}(b_t)}{\bar T_{A_t}(\widehat b_{t,0})}
+
\tv{\bar T_{A_t}(\widehat b_{t,0})}{\bar c_{Z_{t+1}}}
\nonumber\\
&\le
d_{t,0}+\bar\eta_{\mathcal C,0}
\nonumber\\
&\le
\Lambda\bar e_{t,0}+\bar\eta_{\mathcal C,0}.
\end{align*}
Multiplying the recursion by $\gamma^t$ and summing the resulting geometric
series gives
\begin{equation}
\sum_{t=0}^{\infty}\gamma^t\E[\bar e_{t,0}]
\le
\frac{\bar e_{0,\mathcal C}}{1-\gamma\Lambda}
+
\frac{\gamma\bar\eta_{\mathcal C,0}}{(1-\gamma)(1-\gamma\Lambda)}.
\nonumber
\end{equation}
Hence
\begin{equation}
\nonumber
\sum_{t=0}^{\infty}\gamma^t\E[d_{t,0}]
\le
\Lambda
\left(
\frac{\bar e_{0,\mathcal C}}{1-\gamma\Lambda}
+
\frac{\gamma\bar\eta_{\mathcal C,0}}{(1-\gamma)(1-\gamma\Lambda)}
\right),
\end{equation}
which proves the $\delta=0$ bound.
\end{proof}

\section{Additional Experimental Details}
\label{app:app_experimental_details}
\paragraph{Planner oracle.} The truncated planner in Theorem~\ref{thm:recap_regularized_bound} scales exponentially with horizon and observation branching, so we use a PBVI-terminal lookahead oracle instead. In our experiments, PBVI runs for $50$ iterations, and actions are chosen by exact depth $H=3$ lookahead using the PBVI value function as the terminal estimate. Every comparator uses the same oracle.

\paragraph{Pilot cloud.} The pilot cloud provides the candidate set for representatives and must cover the belief space adequately. To this end, we use a mixed sampling strategy. In particular, we sample $N_{\rm pilot}=256$ predictive priors via three strategies: deterministic branch expansion to depth $2$, retaining the top $3$ positive-probability observations and discarding branches below $10^{-8}$ (fraction $0.3$); planner-guided $\epsilon$-greedy rollouts with $\epsilon=0.15$ and horizon $32$ (fraction $0.3$); and random action rollouts (fraction $0.4$). Branch expansion provides near-horizon coverage, guided rollouts concentrate mass on policy-relevant beliefs, and random rollouts fill in regions the greedy policy rarely visits.

\paragraph{Baseline implementations.} All baselines use the same budget $K=2^B$ and the same PBVI-terminal oracle. At runtime, all quantization methods use the same observation-compatible correction and greedy action selection as \textsc{RECAP}. We provide the details for baselines below.

\begin{enumerate}
    \item \textit{Predictive $k$-center} selects $K$ representatives from the pilot cloud by farthest-first greedy in total variation distance. It is the initialization of \textsc{RECAP} without medoid refinement, i.e., $I_{\rm max}=0$.
    
    \item \textit{Predictive $k$-medoids} selects $K$ pilot-cloud points minimizing weighted total variation distortion, initialized by $k$-center and refined by coordinate-sampled swaps. Its objective matches \textsc{RECAP} with $\lambda_{\rm rec}= \lambda_{\rm act}=0$, though the optimizer differs.

    \item \textit{Recursive-feature $k$-medoids} maps each pilot point to a one-step branch-prediction feature vector and runs $k$-medoids in that space, testing whether recursive predictive structure alone captures the main gain without action-relevance or TV-based coverage.

    \item \textit{Value-feature $k$-medoids} maps each pilot point to a normalized vector of branchwise planner $Q$-values, then runs $k$-medoids in that feature space, testing whether action-value proximity alone suffices without TV-based coverage or recursive closure.

    \item \textit{Dirichlet simplex codebook} constructs $K$ representatives from the probability simplex via farthest-first $k$-center on 256 Dirichlet samples, independent of the pilot cloud.

    \item \textit{Finite observation window} computes the longest observation suffix that fits within $K$ states, builds a lookup table from suffixes to empirical average beliefs, and acts greedily from the looked-up belief.
\end{enumerate}
 
\paragraph{Benchmarks.}
Table~\ref{tab:benchmarks} lists the 11 benchmarks with their sizes and budget assignments. All are drawn from~\citep{pomdp_org}.

\begin{table}[h]
\centering
\caption{Benchmark summary. $|\cS|$, $|\cA|$, $|\cY|$: states, actions, observations. $B$: memory budget in bits.}
\label{tab:benchmarks}
\resizebox{\textwidth}{!}{%
\setlength{\tabcolsep}{4pt}
\begin{tabular}{lrrrrl}
\toprule
Benchmark     & $|\cS|$ & $|\cA|$ & $|\cY|$ & $B$ & Brief description \\
\midrule
paint         &  4 &  4 &  2 & 2 & Quality-control: Paint, inspect, ship, or reject a part under noisy blemish signal. \\
marking       &  9 &  4 &  3 & 2 & Robot uses a movable token to mark its environment and reach a goal under partial observation. \\
saci\_s12     & 12 &  6 &  5 & 2 & Aircraft identification: classify friend/enemy from noisy sensor passes. \\
line4\_2goals &  4 &  2 &  1 & 2 & 1D two-goal navigation with no informative observations, agent acts from memory alone. \\
\midrule
shuttle       &  8 &  3 &  5 & 3 & Chrisman docking: shuttle navigation with aliased station/docking observations. \\
network       &  7 &  4 &  2 & 3 & Littman network: regulate server load against crash risk with noisy up-down observations. \\
1d\_noisy     &  4 &  2 &  2 & 3 & Noisy 1D maze: goal observation only; steep discount penalizes slow arrival. \\
query\_s2     &  9 &  2 &  3 & 3 & Route queries to the available resource under noisy response observations. \\
\midrule
mini\_hall2   & 13 &  3 &  9 & 4 & Small hallway navigation with deterministic but ambiguous local observations. \\
4x4           & 16 &  4 &  2 & 4 & 4x4 grid navigation: deterministic moves with only a goal signal, agent acts from memory alone. \\
4x5x2        & 39 &  4 &  4 & 4 & Two-level grid: ambiguous landmark observations require multi-step localization. \\
\bottomrule
\end{tabular}
}
\end{table}


\newpage
\section*{NeurIPS Paper Checklist}
\begin{enumerate}

\item {\bf Claims}
    \item[] Question: Do the main claims made in the abstract and introduction accurately reflect the paper's contributions and scope?
    \item[] Answer: \answerYes{} 
    \item[] Justification: The abstract and introduction accurately reflect the three contributions delivered in the paper by: (i) pointing to main theorems and results; (ii) clearly positioning them against the related literature with explicit distinctions; (iii) stating the tools and theoretical concepts used; (iv) conveying the key insight that control-relevant belief distinctions must survive any lossy memory compression.

\item {\bf Limitations}
    \item[] Question: Does the paper discuss the limitations of the work performed by the authors?
    \item[] Answer: \answerYes{} 
    \item[] Justification: The paper acknowledges that the converse bound depends on the decision witness design. It discusses the point that one should optimize over witnesses to obtain the strongest bound. The paper also discusses \textsc{RECAP}'s dependence on pilot cloud quality in the appendix. 

\item {\bf Theory assumptions and proofs}
    \item[] Question: For each theoretical result, does the paper provide the full set of assumptions and a complete (and correct) proof?
    \item[] Answer: \answerYes{} 
    \item[] Justification: Full proofs with all assumptions, intermediate steps, and supporting propositions appear in the appendix due to space restrictions. Each main-text theorem includes a pointer to its proof. We provide detailed intuitions for each theorem in the main text. All theorems, definitions, and propositions are numbered and cross-referenced. For clarity, displayed equations are numbered only when referenced elsewhere or when they appear as the conclusion of a formal result.

\item {\bf Experimental result reproducibility}
    \item[] Question: Does the paper fully disclose all the information needed to reproduce the main experimental results of the paper to the extent that it affects the main claims and/or conclusions of the paper (regardless of whether the code and data are provided or not)?
    \item[] Answer: \answerYes{} 
    \item[] Justification: The main algorithm, all hyperparameters, evaluation method, baseline implementations, and benchmark configurations are clearly reported in Section~\ref{sec:experimental_setup} and Appendix~\ref{app:app_experimental_details}. The anonymized code repository is available at the link provided in the paper and includes instructions for reproducing all reported results.

\item {\bf Open access to data and code}
    \item[] Question: Does the paper provide open access to the data and code, with sufficient instructions to faithfully reproduce the main experimental results, as described in supplemental material?
    \item[] Answer: \answerYes{} 
    \item[] Justification: The code is publicly available via the anonymized repository linked in the paper. The readme provides exact instructions to reproduce all reported results. The benchmarks used in the paper are publicly available at~\citep{pomdp_org}.

\item {\bf Experimental setting/details}
    \item[] Question: Does the paper specify all the training and test details (e.g., data splits, hyperparameters, how they were chosen, type of optimizer) necessary to understand the results?
    \item[] Answer: \answerYes{} 
    \item[] Justification: All hyperparameters, evaluation method, memory budgets, number of seeds, episode length, and baseline implementations are specified in Section~\ref{sec:experimental_setup} and Appendix~\ref{app:app_experimental_details}.

\item {\bf Experiment statistical significance}
    \item[] Question: Does the paper report error bars suitably and correctly defined or other appropriate information about the statistical significance of the experiments?
    \item[] Answer: \answerYes{} 
    \item[] Justification: We report mean discounted return and standard deviation computed over $3000$ evaluation episodes per method per benchmark, obtained from 3 independent design seeds of 1000 Monte Carlo episodes each. Results are reported across 11 benchmarks spanning diverse POMDP domains and against six budget-matched baselines, ensuring all methods face identical transition and observation randomness, so that performance differences are attributable to the memory design alone.

\item {\bf Experiments compute resources}
    \item[] Question: For each experiment, does the paper provide sufficient information on the computer resources (type of compute workers, memory, time of execution) needed to reproduce the experiments?
    \item[] Answer: \answerYes{} 
    \item[] Justification: The paper provides the details of the computer used for simulations. It states the range of runtime for each benchmark setting and also the runtime required to generate the complete results in Section~\ref{sec:experiments}.
    
\item {\bf Code of ethics}
    \item[] Question: Does the research conducted in the paper conform, in every respect, with the NeurIPS Code of Ethics \url{https://neurips.cc/public/EthicsGuidelines}?
    \item[] Answer: \answerYes{} 
    \item[] Justification: This paper presents foundational theoretical and algorithmic work on POMDP control with no human subjects, no sensitive data, and no foreseeable direct path to harmful applications. The benchmarks are drawn from~\citep{pomdp_org}, which is licensed under Creative Commons Attribution-NonCommercial 4.0 International (CC BY-NC 4.0).

\item {\bf Broader impacts}
    \item[] Question: Does the paper discuss both potential positive societal impacts and negative societal impacts of the work performed?
    \item[] Answer: \answerNA{} 
    \item[] Justification: This is foundational theoretical and algorithmic research on POMDP control with no direct path to harmful applications or deployment.
    
\item {\bf Safeguards}
    \item[] Question: Does the paper describe safeguards that have been put in place for responsible release of data or models that have a high risk for misuse (e.g., pre-trained language models, image generators, or scraped datasets)?
    \item[] Answer: \answerNA{} 
    \item[] Justification: The paper releases only a POMDP control algorithm and benchmark evaluation code, which pose no risk of misuse.

\item {\bf Licenses for existing assets}
    \item[] Question: Are the creators or original owners of assets (e.g., code, data, models), used in the paper, properly credited and are the license and terms of use explicitly mentioned and properly respected?
    \item[] Answer: \answerYes{} 
    \item[] Justification: The only external asset is the POMDP benchmark library~\citep{pomdp_org}, which is properly credited and used under its Creative Commons Attribution-NonCommercial 4.0 International (CC BY-NC 4.0) license.

\item {\bf New assets}
    \item[] Question: Are new assets introduced in the paper well documented and is the documentation provided alongside the assets?
    \item[] Answer: \answerYes{} 
    \item[] Justification: The \textsc{RECAP} codebase is introduced as a new asset. The anonymized GitHub repo is given in the paper, and documented through the algorithm description in Section~\ref{sec:recap_algorithm} and the experimental details in Appendix~\ref{app:app_experimental_details}.

\item {\bf Crowdsourcing and research with human subjects}
    \item[] Question: For crowdsourcing experiments and research with human subjects, does the paper include the full text of instructions given to participants and screenshots, if applicable, as well as details about compensation (if any)? 
    \item[] Answer: \answerNA{} 
    \item[] Justification: The paper does not involve human subjects or crowdsourcing.

\item {\bf Institutional review board (IRB) approvals or equivalent for research with human subjects}
    \item[] Question: Does the paper describe potential risks incurred by study participants, whether such risks were disclosed to the subjects, and whether Institutional Review Board (IRB) approvals (or an equivalent approval/review based on the requirements of your country or institution) were obtained?
    \item[] Answer:  \answerNA{}  
    \item[] Justification: The paper does not involve human subjects or crowdsourcing.

\item {\bf Declaration of LLM usage}
    \item[] Question: Does the paper describe the usage of LLMs if it is an important, original, or non-standard component of the core methods in this research? Note that if the LLM is used only for writing, editing, or formatting purposes and does \emph{not} impact the core methodology, scientific rigor, or originality of the research, declaration is not required.
    \item[] Answer: \answerNA{} 
    \item[] Justification: LLMs were used solely for writing assistance, such as improving conciseness and clarity of exposition.

\end{enumerate}

\end{document}